\documentclass[10pt,twocolumn,letterpaper]{article}

\usepackage{cvpr}
\usepackage{times}
\usepackage{epsfig}
\usepackage{graphicx}
\usepackage{amsmath}
\usepackage{amssymb}
\usepackage{subfig}
\usepackage{authblk}
\usepackage{multirow}
\usepackage{algorithm}
\usepackage{algorithmic}

\usepackage[breaklinks=true,bookmarks=false]{hyperref}
\cvprfinalcopy 
\def\cvprPaperID{0124} 

\ifcvprfinal
\begin{document}

\title{Self-Supervised Adversarial Hashing Networks for Cross-Modal Retrieval}
\author[1]{Chao~Li}
\author[1]{Cheng~Deng\thanks{Corresponding authors}}
\author[1]{Ning~Li}
\author[2]{Wei~Liu$^{*}$}
\author[1]{Xinbo~Gao}
\author[3]{Dacheng~Tao}
\affil[1]{School of Electronic Engineering, Xidian University, Xi'an 710071, China}
\affil[2]{Tencent AI Lab, Shenzhen, China}
\affil[3]{UBTECH Sydney AI Centre, SIT, FEIT, University of Sydney, Australia, 
\authorcr li\_chao@stu.xidian.edu.cn, \{chdeng.xd, ningli2017\}@gmail.com, wliu@ee.columbia.edu, xbgao@mail.xidian.edu.cn, dacheng.tao@sydney.edu.au}
\maketitle
\thispagestyle{empty}
\begin{abstract}
Thanks to the success of deep learning, cross-modal retrieval has made significant progress recently. However, there still remains a crucial bottleneck: how to bridge the modality gap to further enhance the retrieval accuracy. In this paper, we propose a self-supervised adversarial hashing (\textbf{SSAH}) approach, which lies among the early attempts to incorporate adversarial learning into cross-modal hashing in a self-supervised fashion. The primary contribution of this work is that two adversarial networks are leveraged to maximize the semantic correlation and consistency of the representations between different modalities. In addition, we harness a self-supervised semantic network to discover high-level semantic information in the form of multi-label annotations. Such information guides the feature learning process and preserves the modality relationships in both the common semantic space and the Hamming space. Extensive experiments carried out on three benchmark datasets validate that the proposed SSAH surpasses the state-of-the-art methods.
\vspace{-0.5cm}
\end{abstract}

\section{Introduction}
\vspace{-0.15cm}
Owing to the explosive increase in multimedia data from a great variety of search engines and social media, cross-modal retrieval has become a compelling topic in recent years~\cite{liu2011hashing, liu2016query, liu2016structure, liu2016queryhash, liu2017distributed,Mandal2017CVPR, shen2015learning, sun2013survey, Wan2014Deep, Wu2017CVPR, xu2013survey, Zhang2016CVPR}. Cross-modal retrieval aims to search semantically similar instances in one modality (\emph{e.g.}, image) by using a query from another modality (\emph{e.g.}, text). In order to satisfy the requirements of low storage cost and high query speed in real-world applications, hashing has been of considerable interest in the field of cross-modal retrieval, which maps high-dimensional multi-modal data into a common hash code space in such a way that gives similar cross-modal items similar hash codes.
\begin{figure*}[!t]
	\begin{center}
		\includegraphics[width=0.95\textwidth]{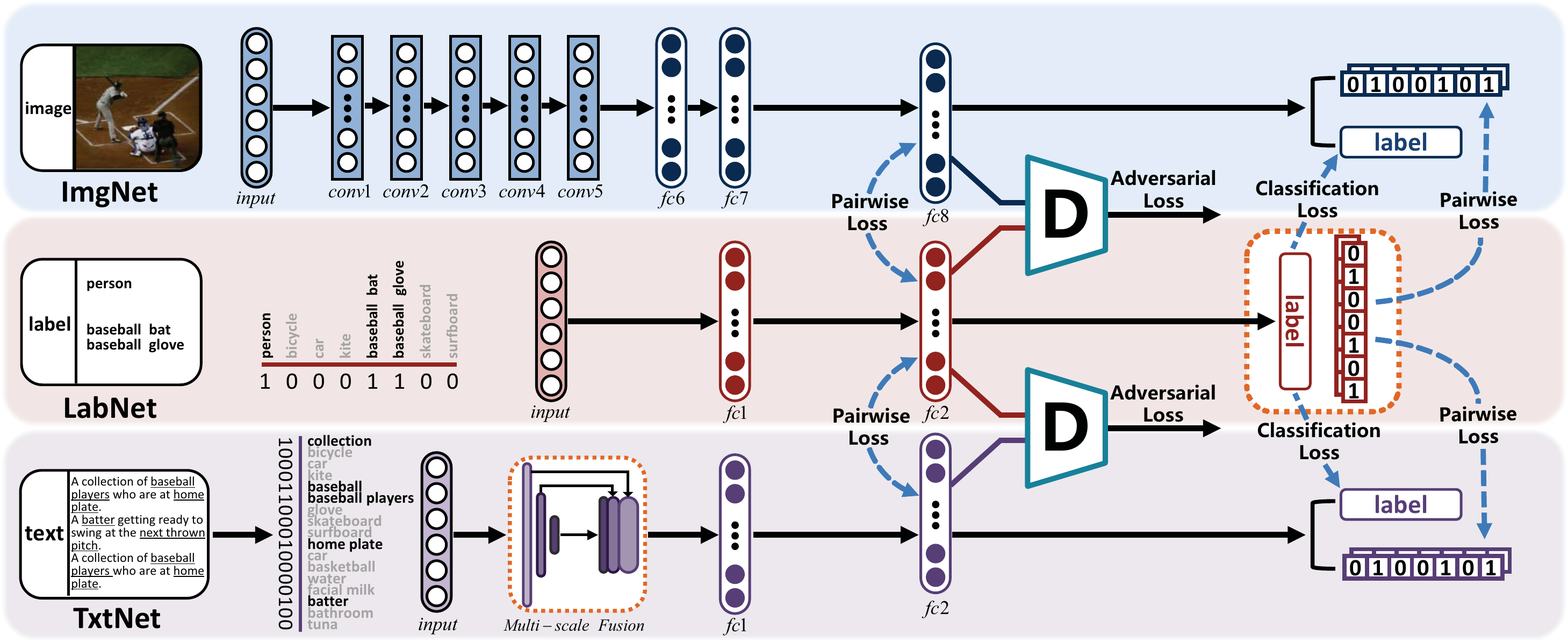}
	\end{center}
	\vspace{-0.6cm}
	\caption{The framework of our proposed SSAH.}
	\label{fig:SSAH framework}
	\vspace{-0.5cm}
\end{figure*}
Since the instances from different modalities are hetrogeneous in terms of their feature representation and distribution, \emph{i.e.}, their modality gap, it is necessary to explore their semantic relevance in sufficient detail to bridge this modality gap. Most existing shallow cross-modal hashing methods (in both unsupervised~\cite{Bronstein2010Data, Feng2014Cross, Kumar2011Learning, liu2014discrete} and supervised settings~\cite{deng2016discriminative, Liu2017CVPR, liu2012supervised, Masci2014Multimodal, shen2015supervised, Wu2014Sparse, song2015top}), always capture the semantic relevance in a common Hamming space. Compared with  their unsupervised counterparts, supervised cross-modal hashing methods can achieve superior performance by exploiting semantic labels or information concerning relevance, thereby distilling a cross-modal correlation. However, almost all these existing shallow cross-modal hashing methods are based on hand-crafted features, which may limit the discriminative representation of instances and thus degrade the accuracy of the learned binary hash codes.

\vspace{-0.05cm}
In recent years, deep learning has become very successful at learning highly discriminative features for various applications~\cite{bengio2013representation}\cite{krizhevsky2012imagenet}. However, only a few works have performed deep learning for cross-modal hashing~\cite{Cao2016Deep, Liong2017ICCV, Jiang2017CVPR, Shen2017ICCV, yang2017pairwise}, which can capture nonlinear correlations among cross-modal instances more effectively. It is worth noting that there are still some common disadvantages hindering the current deep cross-modal hashing methods. First, these methods simply and directly adopt single-class labels to measure the semantic relevance across modalities~\cite{Liong2017ICCV}\cite{Jiang2017CVPR}. In fact, in standard cross-modal benchmark datasets such as NUS-WIDE~\cite{chua2009nus} and Microsoft COCO~\cite{lin2014microsoft}, an image instance can be assigned to multiple category labels~\cite{Ranjan2015ICCV}, which is beneficial as it permits semantic relevance to be described more accurately across different modalities. Second, these methods enforce a narrowing of the modality gap by constraining the corresponding hash codes with certain pre-defined loss functions~\cite{Cao2016Correlation}. The code length is usually less than 128 bits. This means that most of the useful information is neutralized, making the hash codes incapable of capturing the inherent modality consistency. In comparison, high-dimensional modality-specific features contain more abundant information that helps to bridge the modality gap. Therefore, how to encourage more abundant semantic relevance and build more accurate modality relationships becomes crucial to achieve satisfactory performance in real-world retrieval applications.

In this paper, we propose a novel self-supervised adversarial hashing (SSAH) method to aid in cross-modal retrieval. Specifically, we employ two adversarial networks to jointly learn the high-dimensional features and their corresponding hash codes for different modalities. At the same time, a self-supervised semantic network is seamlessly integrated to discover semantic information in the form of multi-label annotations, with which the adversarial learning is supervised to maximize the semantic relevance and the feature distribution consistency between modalities. The highlights of our work can be outlined as follows:
\begin{itemize}
	\vspace{-0.1cm}
	\item We propose a novel self-supervised adversarial hashing method for cross-modal retrieval. As far as we know, this is one of the first attempts to utilize adversarial learning in an aim to tackle the cross-modal hashing problem.
	\vspace{-0.15cm}
	\item We integrate self-supervised semantic learning with adversarial learning in order to preserve the semantic relevance and the representation consistency across modalities as much as possible. In this way, we can effectively bridge the modality gap.
	\vspace{-0.15cm}
	\item Extensive experiments conducted on three benchmark datasets demonstrate that our proposed SSAH significantly outperforms the  current state-of-the-art cross-modal hashing methods, including both traditional and deep-learning-based methods.
	\vspace{-0.05cm}
\end{itemize}

The rest of this paper is organized as follows. Related work in cross-modal hashing is introduced in Section~\ref{section: Related Work}. Our proposed SSAH model and the learning algorithm are presented in Section~\ref{section: Proposed SSAH}. Experiments are shown in Section~\ref{section: Experiment} and Section~\ref{section: Conclusion} concludes this work.

\vspace{-0.25cm}
\section{Related Work}
\vspace{-0.15cm}
\label{section: Related Work}
Cross-modal hashing methods can be roughly categorized into unsupervised and supervised settings. Unsupervised hashing methods~\cite{Ding2014CVPR, Song2013Inter, Wang2015Semantic, zhou2014latent} learn hashing functions by discovering the inter-modality and intra-modality information belonging to the unlabeled training data. Ding~\emph{et al.}~\cite{Ding2014CVPR} learned a unified binary code by performing a matrix factorization with a latent factor model. The work of Song~\emph{et al.}~\cite{Song2013Inter} learns functions that can map features from different modalities into the common Hamming space.

Supervised hashing methods~\cite{Bronstein2010Data, Cao2016Correlation, Kumar2011Learning, Lin2015CVPR, wang2013learning, Wu2014Sparse, Zhang2014Large} aim to exploit available supervised information (such as labels or the semantic affinities of training data) to improve performance. Brostein \emph{et al.}~\cite{Bronstein2010Data} present a cross-modal hashing approach by preserving the intra-class similarity via eigen-decomposition and boosting. Semantic correlation maximization (SCM)~\cite{Zhang2014Large} utilizes label information to learn a modality-specific transformation, which preserves the maximal correlation between modalities. Semantics-preserving hashing (SePH)~\cite{Lin2015CVPR} generates a unified binary code by modeling an affinity matrix in a probability distribution while at the same time minimizing the Kullback-Leibler divergence. Most of these methods depend on hand-crafted features that have to be extracted by shallow architectures; as such, these methods make it difficult to effectively exploit the heterogeneous relationships across modalities.

Recently, some works have reported on deep cross-modal hashing retrieval~\cite{Cao2016Deep, Liong2017ICCV, Jiang2017CVPR, Wang2017MM}. Deep cross-modal hashing (DCMH)~\cite{Jiang2017CVPR} performs an end-to-end learning framework, using a negative log-likelihood loss to preserve the cross-modal similarities. Adversarial cross-modal retrieval (ACMR)~\cite{Wang2017MM} directly borrows from the adversarial-learning approach and tries to discriminate between different modalities using a classification manner that is the one most closely related to ours. In comparison to~\cite{Wang2017MM}, however, our SSAH utilizes two adversarial networks to jointly model different modalities and thereby further capture their semantic relevance and representation consistence under the supervision of the learned semantic feature.
\vspace{-0.3cm}
\section{Proposed SSAH}
\vspace{-0.15cm}
\label{section: Proposed SSAH}
Without loss of generality, we focus on cross-modal retrieval for bimodal data (\emph{i.e.}, image and text). Fig.~\ref{fig:SSAH framework} is a flowchart showing the general principles of the proposed SSAH method. This method  mainly consists of three parts, including a self-supervised semantic generation network called \emph{LabNet}, and two adversarial networks called \emph{ImgNet} and \emph{TexNet} for image and text modalities, respectively.

Specifically, the target of \emph{LabNet} is framed in a way that allows it to learn semantic features from multi-label annotations. It can then be regarded as a common semantic space in which to supervise modality-feature learning over two phases. In the first phase, modality-specific features from separate generator networks are associated with each other in a common semantic space. Since each output layer in a deep neural network contains semantic information, associating modality-specific features in a common semantic space can help to promote the semantic relevance between modalities. In the second phase, semantic features and modality-specific features are simultaneously fed into two discriminator networks. As a result, the feature distributions of the two modalities tend to become consistent under the supervision of the same semantic feature. In this section, we present the details about our SSAH method, including the methods behind the model formulation and the learning algorithm.
\vspace{-0.2cm}
\subsection{Problem Formulation}
\vspace{-0.15cm}
\label{section: Problem Definition}
Let $O=\{o_{i}\}_{i=1}^{n}$ denote a cross-modal dataset with $n$ instances, $o_{i}=(v_{i}, t_{i}, l_{i})$, where $v_{i}\in\mathbb{R}^{1\times d_{v}}$ and $t_{i}\in\mathbb{R}^{1\times d_{t}}$ are the original image and text features for the $i$-th instance, and $l_{i}=[l_{i1},\ldots,l_{ic}]$ is the multi-label annotation assigned to $o_{i}$, where $c$ is the class number. If $o_{i}$ belongs to the $j$-th class $l_{ij}=1$, otherwise $l_{ij}=0$. The image-feature matrix is defined as $V$, the text-feature matrix as $T$, and the label matrix as $L$ for all instances. The pairwise multi-label similarity matrix $S$ is used to describe semantic similarities between each of the two instances, where $S_{ij}=1$ means that $o_{i}$ is semantically similar to $o_{j}$, otherwise $S_{ij}=0$. In a multi-label setting, two instances ($o_{i}$ and $o_{j}$) are annotated by multiple labels. Thus, we define $S_{ij} = 1$ if $o_{i}$ and $o_{j}$ share as least one label, otherwise $S_{ij}=0$.

The goal of cross-modal hashing is to learn a unified hash code for the two modalities: $B^{v,t} \in \{-1,1\}^{K}$, where $K$ is the length of the binary code. The similarity between two binary codes is evaluated using the Hamming distance. The relationship between their Hamming distance $dis_{H}(b_{i},b_{j})$ and their inner product $\left<b_{i},b{_j}\right>$ can be formulated using $dis_{H}(b_{i},b_{j}) = \frac{1}{2}(K-\left<b_{i},b{_j}\right>)$.
So, we can use the inner product to quantify the similarity of two binary codes. Given $S$, the probability of $S$ under the condition $B$ can be expressed as:
\vspace{-0.2cm}
\begin{equation} \label{eq:loglikehood}
\begin{aligned}
p\left( { S }_{ ij }| B \right) =\begin{cases} \delta \left( { \Psi  }_{ ij } \right) \text{,}& { S }_{ ij }=1 \\ 1-\delta \left( { \Psi  }_{ ij } \right) \text{,}& { S }_{ ij }=0 \end{cases}
\end{aligned}
\vspace{-0.1cm}
\end{equation}
where $\delta \left( {\Psi}_{ ij } \right)=\frac{1}{1+e^{-{\Psi }_{ ij }}}$, and ${\Psi}_{ij}=\frac{1}{2}\left<b_{i},b{_j}\right>$. Therefore, two instances with a larger inner product should be similar with a high probability. The problem of quantifying the similarity between binary codes in the Hamming space can thereby be transformed into a calculation of the inner product of the codes' original features.

Here, we frame a couple of adversarial networks (\emph{ImgNet} and \emph{TxtNet}) to learn separate hash functions for image and text modalities (\emph{i.e.,} $H^{v,t}=f^{v,t}(v,t;\theta^{v,t})$). At the same time, we construct an end-to-end self-supervised semantic network (\emph{LabNet}) in order to model the semantic relevance between image and text modality in a common semantic space while learning the hash function for the semantic feature (\emph{i.e.,} $H^{l}=f^{l}(l;\theta^{l})$). Here, $f^{v,t,l}$ are hash functions, and $\theta^{v,t,l}$ are the network parameters to be learned. With the learned $H^{v,t,l}$, binary codes $B^{v,t,l}$ can be generated by applying a sign function to $H^{v,t,l}$:
\vspace{-0.1cm}
\begin{equation} \label{eq: hashing code}
\begin{aligned}
\resizebox{0.7\hsize}{!}{$B^{v,t,l} \ = \ sign(H^{v,t,l})\ \in \ \{-1,1\}^{K}$}\\
\end{aligned}
\vspace{-0.1cm}
\end{equation}
To make this easier to understand, we additionally use $F^{v,t,l} \in \mathbb{R}^{s\times n}$ to denote the semantic features in a common semantic space for images, text and labels, $s$ is the dimension of the semantic space. In practice, $F^{v,t,l}$ correspond to certain output layers of deep neural networks (\emph{ImgNet}, \emph{TxtNet} and \emph{LabNet}, respectively).
\vspace{-0.1cm}
\subsection{Self-supervised Semantic Generation}
\vspace{-0.15cm}
Taking the Microsoft COCO dataset as an example, there is an instance that is annotated with multiple labels, such as ``person'', ``baseball bat'' and ``baseball glove''. In this scenario, the most natural thought is that it is possible to take the multi-label annotation as a conduciveness with which to bridge the semantic relevance between modalities at a more fine-grained level. 
We have designed an end-to-end full-connected deep neural network, named \emph{LabNet}, to model semantic relevance between different modalities. Given a multi-label vector for an instance, \emph{LabNet} extracts abstract semantic features layer by layer; with these we can supervise the feature-learning process in both \emph{ImgNet} and \emph{TxtNet}. Since a triplet $(v_i, t_i, l_i)$ is used to describe the same $i$-th instance, we regard $l_i$ as self-supervised semantic information for $v_i$ and $t_i$. In~\emph{LabNet}, semantic features are projected into their corresponding hash codes through nonlinear transformation. Our intention is that the similarity relationships between semantic features and their corresponding hash codes is well preserved; this is the fundamental premise behind the efficient association between different modalities. Accordingly, for \emph{LabNet}, the final objective can be formulated as follows:
\vspace{-0.15cm}
\begin{equation}
\label{eq:labobjFunc}
\begin{aligned}
\min\limits_{B^{l},\theta^{l}, \hat{L}^{l}}\mathcal{L}^{l} \ =& \alpha \mathcal{J}_{1} + \gamma \mathcal{J}_{2} + \eta \mathcal{J}_{3} + \beta \mathcal{J}_{4} \\
=&\resizebox{0.7\hsize}{!}{$-\alpha \sum _{ i,j=1 }^{ n }{ \left( { S }_{ ij }{ \Delta  }_{ ij }^{l}-\log { \left( 1+{ e }^{ { \Delta  }_{ ij }^{l} } \right)  }  \right)  }$}  \\
&\resizebox{0.7\hsize}{!}{$-\gamma \sum _{ i,j=1 }^{ n }{ \left( { S }_{ ij }{ \Gamma  }_{ ij }^{l}-\log { \left( 1+{ e }^{ { \Gamma  }_{ ij }^{l} } \right)  }  \right)  }$} \\
&\resizebox{0.6\hsize}{!}{$+\eta { \left\| {H}^{l}-{ B }^{ l } \right\|  }_{ F }^{ 2 }+\beta { \left\| { \hat{ L } }^{l}- L \right\|  }_{ F }^{ 2 }$}\\
&\emph{s.t.} \quad {B}^{l}\in \{-1, 1\}^K
\end{aligned}
\end{equation}
where ${\Delta}_{ij}^{l}=\frac{1}{2}({F_{\ast i}^{l} })^{\top}({F}_{\ast j}^{l})$, ${\Gamma }_{ij}^{l}=\frac{1}{2}({H_{\ast i}^{l}})^{\top}({ H_{\ast j}^{l} })$, $H^{l}$ are predicted hash codes and $\hat{L}^{l}$ are predicted labels. $\alpha$, $\gamma$, $\eta$ and $\beta$ are hyper-parameters. In~\eqref{eq:labobjFunc}, $\mathcal{J}_{1}$ and $\mathcal{J}_{2}$ are two negative-log likelihood functions. $\mathcal{J}_{1}$ is used to preserve the similarity between semantic features, and $\mathcal{J}_{2}$ is used to preserve the instances where the similar label information has similar hash codes. $\mathcal{J}_{3}$ is the approximation loss for the binarization of the learned hash codes, and $\mathcal{J}_{4}$ is the classification loss of the original label and the predicted label.
\vspace{-0.4cm}
\subsection{Feature Learning}
\vspace{-0.2cm}
As described above, the different modalities belonging to a multi-modal instance are semantically relevant. In order to preserve this semantic relevance, we supervise the feature-learning process for two modalities under~\emph{LabNet}'s guidance, including  the supervision of the semantic features and the learned binary codes. To address image modality, we have designed an end-to-end feature-learning network, named~\emph{ImgNet}, which can project images into hash codes. By supervising the image-feature learning using the semantic network, we can keep the same semantic relevance between~\emph{ImgNet} and the semantic network. This is the self-supervised role of the semantic network when used in~\emph{ImgNet}. Similarly, when considering text modality, we use the semantic network to supervise the feature-learning process of~\emph{TxtNet} in the same way. Thus, the objective function of self-supervised feature learning for different modalities in $v$ and $t$ can be written as:
\vspace{-0.1cm}
\begin{equation} \label{eq:imgobjFunc}
\begin{aligned}
\min\limits_{B^{v,t},\theta^{v,t}}\mathcal{L}^{v,t} =&\alpha \mathcal{J}_{1} + \gamma \mathcal{J}_{2} + \eta \mathcal{J}_{3} + \beta \mathcal{J}_{4} \\
=&\resizebox{0.7\hsize}{!}{$-\alpha \sum _{ i,j=1 }^{ n }{ \left( { S }_{ ij } { \Delta  }_{ ij }^{ v,t  }-\log { \left( 1+{ e }^{ { \Delta  }_{ ij }^{v,t} } \right)} \right)  } $}\\
&\resizebox{0.7\hsize}{!}{$-\gamma \sum _{ i,j=1 }^{ n }{ \left( { S }_{ ij }{ \Gamma  }_{ ij }^{v,t}-\log { \left( 1 +{ e }^{ { \Gamma  }_{ ij }^{v,t} } \right)  }  \right)  } $}\\
&\resizebox{0.6\hsize}{!}{$+\eta { \left\| {H^{v,t}}-{B}^{v,t} \right\| }_{ F }^{2}
+\beta { \left\| { \hat{ L } }^{v,t}- L \right\|  }_{ F }^{2} $}\\
&\emph{s.t.} \quad {B}^{v,t}\in \{-1, 1\}^K
\end{aligned}
\end{equation}
where ${\Delta }_{ij}^{v,t}=\frac{1}{2}({F_{\ast i}^{l}})^{\top}({F_{\ast j}^{v,t}})$, and ${\Gamma }_{ij}^{ v,t}=\frac{1}{2} ({H_{\ast i}^{l}})^{\top} (H_{\ast j}^{v,t})$. $H^{v,t}$ are predicted hash codes and $\hat{L}^{v,t}$ are predicted labels for images and text, respectively. $\alpha$, $\gamma$, $\eta$ and $\beta$ are hyper-parameters. $\mathcal{J}_{1}$ and $\mathcal{J}_{2}$ are two negative-log likelihood functions. $\mathcal{J}_{3}$ and $\mathcal{J}_{4}$ are approximation loss and classification loss defined in a way that is similar to that used in~\emph{LabNet}. It should be noted that although~\eqref{eq:labobjFunc} and ~\eqref{eq:imgobjFunc} are similar in structure they have different meanings. As such, we use the supervised information ${F_{\ast i}^{l}}$ and ${H_{\ast i}^{l}}$ (learned from the semantic network) to guide the process of learning in~\emph{ImgNet} and~\emph{TxtNet}. The relevance can be established using the semantic network. As a result, the modality gap can then be alleviated.

In comparison to image modality, an instance in text modality, generally represented by a bag-of-words (BoW) vector, easily results in sparsity. Therefore, BoW is unsuitable when aiming to discover valuable information needed for learning hash codes. To solve the problem, we have designed a multi-scale fusion model, which consists of multiple average pooling layers and a $1\times1$ convolutional layer. Multiple average pooling layers are used to extract multiple scale features for text data, following which the $1\times1$ convolutional layer is used to fuse multiple features. Through this process, the correlation between different words can also be captured, which is useful when building semantic relevance for text modality. More detailed parameter information is given in Section~\ref{section:Implementation Details}.
\vspace{-0.2cm}
\subsection{Adversarial Learning}
\vspace{-0.15cm}
Under the supervision of~\emph{LabNet}, the semantic relevance can be preserved across different modalities. However, different modalities usually are inconsistently distributed, which is not beneficial if we want to generate unified hash codes. In order to bridge this modality gap and enable more accurate retrieval, we have studied the distribution agreement for different modalities in an adversarial learning manner. We have built two discriminators for image and text modalities to discover their distribution differences. For the image (text) discriminator, the inputs are image (text) modality features and semantic features generated through~\emph{LabNet}, and the output is one single value, either ``0'' or ``1''. Specifically, we define the modality label for the semantic feature that has been generated from a label as ``1'' and define the modality label for image (text) semantic modality features generated from~\emph{ImgNet} (\emph{TxtNet}) as ``0''. We feed $F^{v}$ and $F^{l}$ into the discriminator that has been designed for images and feed $F^{t}$ and $F^{l}$ into another discriminator that has been designed for text. To formulate this structure, let $Y=\{y_{i}\}_{i=1}^{3\times n}$, $y_{i}\in \{ 0, 1\}$ denote the modality label assigned to the semantic feature in the shared common space. Let $Y^{l}=\{y_{i}^{l}\}_{i=1}^{n}$, $y_{i}^{l}=1$ denote the modality labels for the label. Let $Y^{v,t}=\{y_{i}^{v,t}\}_{i=1}^{n}$ and $y_{i}^{v,t}=0$ denote the modality labels for image and text, respectively. When training our model, these two discriminators act as the two adversaries. As such, the objective function can be written as follows:
\vspace{-0.25cm}
\begin{equation}
\label{eq:advobjFunc}
\min\limits_{\theta_{adv}^{\star,l}}\mathcal{L}_{adv}^{\star,l} =  \sum _{ i=1 }^{ 2\times n }{ { {\left\| { D^{\star,l}({ x }_{ i }^{ \star l })}-{y_{i}}^{\star,l} \right\|  }_{2}^{2} } } \ , \ \star = v,t
\vspace{-0.25cm}
\end{equation}
where $x_{i}^{v,t,l}$ is the semantic feature in the common semantic space, while the modality label is $y_{i}^{v,t,l}$, $2\times n$, denoting the number of instances that are fed into each discriminator. The result of~\eqref{eq:advobjFunc} is that the discriminators act as two binary classifiers, classifying the input semantic feature into class ``1'' and class ``0''.
\begin{algorithm}[!t]
	\algsetup{linenosize=\tiny} \scriptsize
	\caption{Pseudopod showing the optimization of our SSAH}
	\label{alg::algorithm}
	\begin{algorithmic}
		\REQUIRE Image set $V$ ; Text set $T$ ; Label set $L$ ;
		\ENSURE Optimal code matrix $B$
		\STATE \textbf{Initialization} \STATE Initialize parameters: $\theta ^{v,t,l}$, $\theta_{adv}^{v,t}$, $\alpha$, $\gamma$, $\eta$, $\beta$\\ 
		\STATE learnrate: $\mu$, mini-batch size: $N^{v,t,l}=128$, maximum iteration number: $T_{max}$.
		\REPEAT
		\FOR{$t$ iteration}
		\STATE Update $\theta ^{l}$ by BP algorithm:
		\STATE ${ \theta  }^{ l }\leftarrow { \theta  }^{ l }-\mu \cdot { \nabla  }_{ { \theta  }^{l} }\frac { 1 }{ n } \left( { \mathcal{L} }_{gen}-{ \mathcal{L} }_{adv} \right)$
		\STATE Update the parameter $\theta ^{v,t}$ by BP algorithm:
		\STATE ${ \theta  }^{\star}\leftarrow { \theta  }^{\star}-\mu \cdot { \nabla  }_{ { \theta  }^{\star} }\frac { 1 }{ n } \left( { \mathcal{L} }_{ gen }-{ \mathcal{L} }_{ adv } \right), \star = v,t$
		\STATE Update ${ \theta}_{adv}^{\star}$ by BP algorithm:
		\STATE ${\theta}^{\star}_{adv}\leftarrow { \theta  }^{\star}_{adv}-\mu \cdot { \nabla  }_{ { \theta  }^{\star}_{adv} }\frac { 1 }{ n } \left( { \mathcal{L} }_{ (gen) }-{ \mathcal{L} }_{adv} \right) , \star = v,t$
		\ENDFOR
		\STATE Update the parameter $B$ by \\
		$B = sign(H+F+G)$
		\UNTIL{convergence}
	\end{algorithmic}
\end{algorithm}
\vspace{-0.25cm}
\subsection{Optimization}
\vspace{-0.15cm}
It is noted that three kinds of hash codes can be generated using our SSAH: $B^{v,t,l} = sign(H^{v,t,l})$. During the training process, we make $B=sign(H^{v}+H^{t}+H^{l})$ to train our model to generate similar binary codes for semantically similar instances. As mentioned above, the overall objective function can be written as follows:
\vspace{-0.2cm}
\begin{equation}
\label{eq:objFuncDivide}
\begin{aligned}
\mathcal{L}_{gen} \ =& \ \mathcal{L}^{v} +\mathcal{L}^{t} + \mathcal{L}^{l} \\
\mathcal{L}_{adv} \ =& \  \mathcal{L}_{adv}^{v} + \mathcal{L}_{adv}^{t} \\
\end{aligned}
\vspace{-0.2cm}
\end{equation}
\vspace{-0.2cm}
If we put them together, we can obtain:
\vspace{-0.05cm}
\begin{equation}
\label{eq:objFunc}
\begin{aligned}
( B, \theta^{v,t,l}) \ = \ & \underset {B, \theta^{v,t,l}}{ arg min } \ \mathcal{L}_{gen}(B, \theta^{v,t,l})-\mathcal{L}_{adv}(\hat{\theta}_{adv}) \\
\theta_{adv} \ = \ & \underset { \theta_{adv} } { \ arg max \ }\mathcal{L}_{gen}(\hat{B}, \hat{\theta}^{v,t,l})- \mathcal{L}_{adv}(\theta_{adv}) \\
\emph{s.t.}& \quad  B \in \{-1, 1\}^K
\end{aligned}
\vspace{-0.1cm}
\end{equation}

Due to the discreteness of parameter $B$ and the vanishing-gradient problem caused by the minimax loss, the optimization of~\eqref{eq:objFunc} is intractable. Hence, we optimize the objective~\eqref{eq:objFunc} through iterative optimization. Firstly, we optimize the $\mathcal{L}^{l}$ over $\theta^{l}$, $B^{l}$, and $\hat{L}^{l}$ by exploring label information. Then, we optimize $\mathcal{L}^{v}$ over $\theta^{v}$ and $B^{v}$ by fixing $\theta^{l}$ and $B^{l}$. Similarly, we leave $\theta^{l}$ and $B^{l}$ fixed to learn $\theta^{t}$ and $B^{t}$, allowing the optimization of $\mathcal{L}^{t}$. During this process, two kinds of modality features are learned in a self-supervised learning manner. Finally, we optimize $\mathcal{L}_{adv}^{v,t}$ over $\theta^{v,t}$ by fixing $\theta^{v,t,l}$. It is noted that all network parameters are learned by utilizing the stochastic gradient descent (SGD) with the back-propagation (BP) algorithm, which is widely adopted in existing deep-learning methods. Algorithm~\ref{alg::algorithm} outlines the whole learning algorithm in detail.

As for out-of-sample extensions: the proposed framework can be applied to cross-modalities. Indeed, it is not limited to two modalities; rather, it can easily be adapted to solve the problems in situations with more than two modalities. Hash codes for the unseen data-point, which may come from different modalities, images or text, can be directly obtained by feeding the original feature into our model:
\begin{equation}
\label{eq:generateImgHash}
\begin{aligned}
b_{q}^{ v,t,l} = sign(f^{ v,t,l}(b_{q}; \theta^{ v,t,l}))\
\end{aligned}
\vspace{-0.1cm}
\end{equation}
Moreover, by feeding the label information into $LabNet$ we can obtain hash codes for the label information, which can then be used to retrieve the related results from both images and text at same time.
\vspace{-0.2cm}
\subsection{Implementation Details}
\vspace{-0.15cm}
\label{section:Implementation Details}
\textbf{Self-Supervised Semantic Network:} We built~\emph{LabNet} with four-layer feed-forward neural networks, which are used to project a label into hash codes ($L\rightarrow 4096 \rightarrow 512 \rightarrow N$). The nodes of output layer $N$ are related to the length of the hash code $K$ and the total class labels $c$ for different datasets, $N=K+c$.

\textbf{Generative Network for Images:} We built~\emph{ImgNet} based on CNN-F \cite{chatfield2014return} neural networks. In order to apply CNN to our SSAH model, we reserve the first seven layers (which were the same as those in CNN-F). Following this, a middle layer fc8 (with 512 nodes) and final output layer (with $N$ nodes) are framed. In addition, we also evaluated our method using the vgg19~\cite{simonyan2014very} network; here, we replaced the CNN-F network with the vgg19 network and left the rest remain unchanged.

\textbf{Generative Network for Text:} We built \emph{TxtNet} using a three-layer feed-forward neural network and a multi-scale (MS) fusion model (T$\rightarrow$MS$\rightarrow$4096$\rightarrow$512$\rightarrow$N). MS consists of a five-level pooling layer (1$\times$1, 2$\times$2, 3$\times$3, 5$\times$5, and 10$\times$10).

\textbf{Adversarial Networks:} We built the discriminator networks using a three-layer feed-forward neural network ($F^{v,t,l}$ $\rightarrow$ 4096 $\rightarrow$ 4096 $\rightarrow$ 1).

Regarding the activate function used in SSAH: $sigmoid$ activation is used to output the predicted label; $tanh$ activation is used to output the hash codes; and the rest of the layers are all uniformly activated by the $relu$ function. In addition, SSAH is implemented via \emph\textbf{TensorFlow} and is run on a server with two NVIDIA TITAN X GPUs.

\vspace{-0.3cm}
\section{Experiment}
\vspace{-0.2cm}
\label{section: Experiment}
\subsection{Datasets}
\vspace{-0.15cm}
The~{\emph{\textbf{MIRFLICKR-25K}}} dataset \cite{huiskes2008mir} contains 25,000 instances collected from Flickr. Each image is labeled with its associated textual tags. Here, we follow the experimental protocols given in DCMH \cite{Jiang2017CVPR}. In total, 20,015 data points have been selected for our experiment. The text for each point is represented as a 1,386-dimensional BoW vector, and each point is manually annotated with at least one of the 24 unique labels.

The~\emph{\textbf{NUS-WIDE}} dataset~\cite{chua2009nus} is a public web image data-set containing 269,648 web images. There are 81 ground-truth concepts that have been manually annotated for search evaluation. After pruning the data that is without any label or tag information, a subset of 190,421 image-text pairs that belong to some of the 21 most-frequent concepts are selected to serve as our dataset.

The~\emph{\textbf{MS COCO}} dataset~\cite{lin2014microsoft} contains about 80,000 training images and 40,000 validation images. Five thousand images from the validation set are selected randomly. In total, there are 85,000 data items have been  used in our experiment. Each data item consists of one image-text pair for two different modalities, and each text is represented as a 2,000-dimension BoW vector. Table~\ref{datasets: all} summarizes the statistics of the three datasets.
\vspace{-0.25cm}
\subsection{Evaluation and Baselines}
\vspace{-0.15cm}
\textbf{Evaluation:} The Hamming ranking and hash lookup are two classical retrieval protocols used to evaluate the performance of a cross-modal retrieval task. In our experiments, we use three evaluation criteria: mean average precision (MAP), which is used to measure the accuracy of the Hamming distances; the precision-recall (PR) curve, which is used to measure the accuracy of the hash lookup protocol; and the precision at n (P\emph{@}n) curve used to evaluate precision by considering only the number of top returned points.

\textbf{Baselines:} We compare our SSAH using six state-of-the-art methods, including  several shallow-structure-based methods (CVH~\cite{Kumar2011Learning}, STMH~\cite{Wang2015Semantic}, CMSSH~\cite{Bronstein2010Data}, SCM \cite{Zhang2014Large}, SePH \cite{Lin2015CVPR}), and a deep-structure-based method (DCMH \cite{Jiang2017CVPR}). In order to conduct a fair comparison, we utilize both CNN-F~\cite{chatfield2014return} and vgg19~\cite{simonyan2014very}, which have both been pre-trained on the ImageNet datasets~\cite{russakovsky2015imagenet} in order to extract deep features for all shallow-structure-based baselines.
\begin{table}[!t]
	\centering
	\caption{Statistics of the datasets used in our experiments.}
	\vspace{-0.25cm}
	\label{datasets: all}
	\begin{tabular}{|l|c|c|c|}
		\hline
		Dataset         & Total      & Train  / Test       & Labels\\
		\hline
		MIRFLICKR-25K   &   20,015   &    10,000 / 2,000    & 24     \\
		\hline
		NUS-WIDE        &  190,421   &    10,500 / 2,100    & 21     \\
		\hline
		MS COCO  &   85,000          &    10,000 / 5,000    & 80     \\
		\hline
	\end{tabular}
	\vspace{-0.7cm}
\end{table}

\begin{figure*}[!t]
	\centering
	\begin{minipage}[b]{0.24\textwidth}
		\vspace{-0.1cm}
		\centering
		\includegraphics[width=0.95\textwidth]{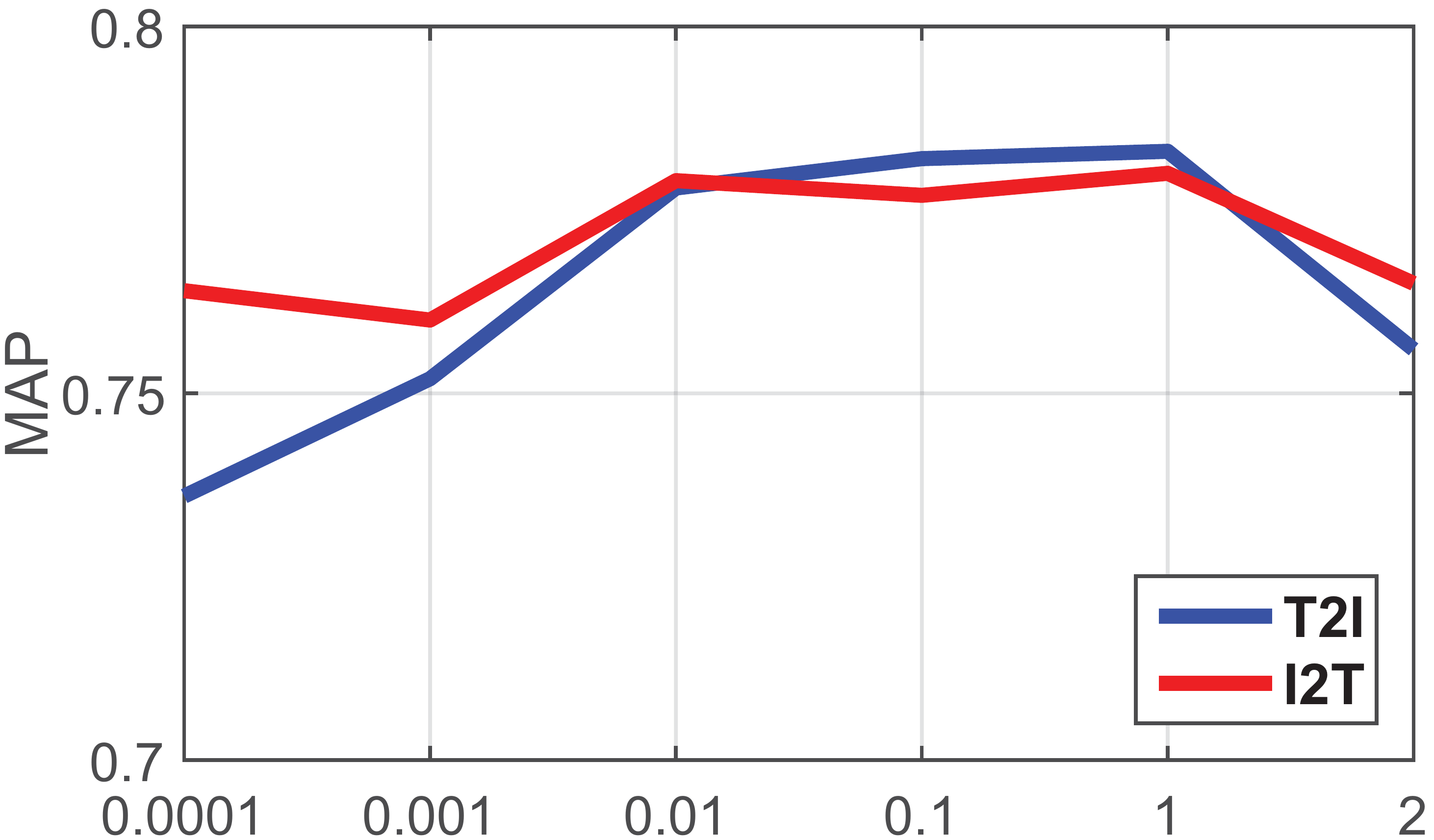}
		\vspace{-0.3cm}
		\caption*{(a) $\alpha$}
	\end{minipage}%
	\begin{minipage}[b]{0.24\textwidth}
		\vspace{-0.25cm}
		\centering
		\includegraphics[width=0.95\textwidth]{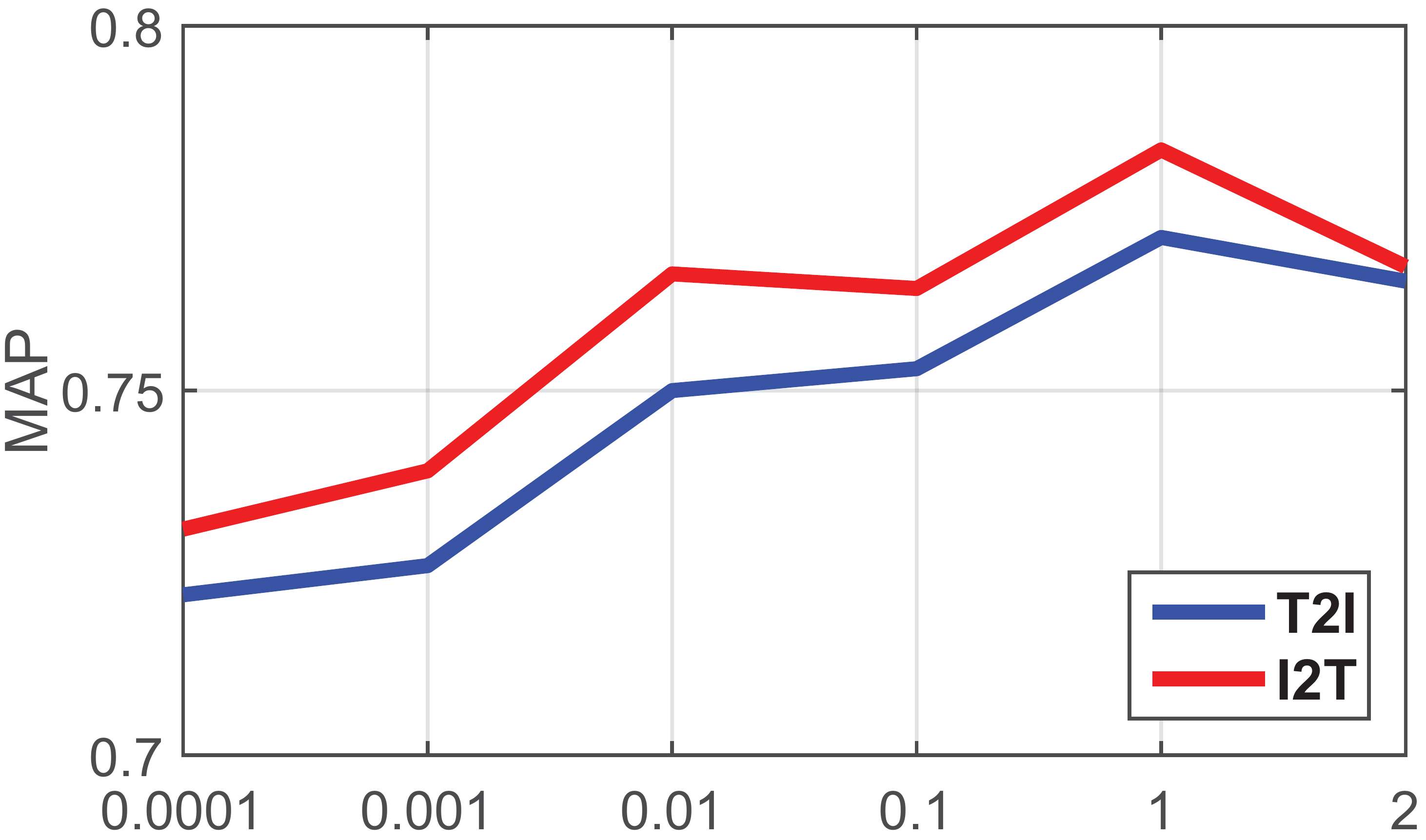}
		\vspace{-0.3cm}
		\caption*{(b) $\gamma$}
	\end{minipage}
	\begin{minipage}[b]{0.24\textwidth}
		\vspace{-0.25cm}
		\centering
		\includegraphics[width=0.95\textwidth]{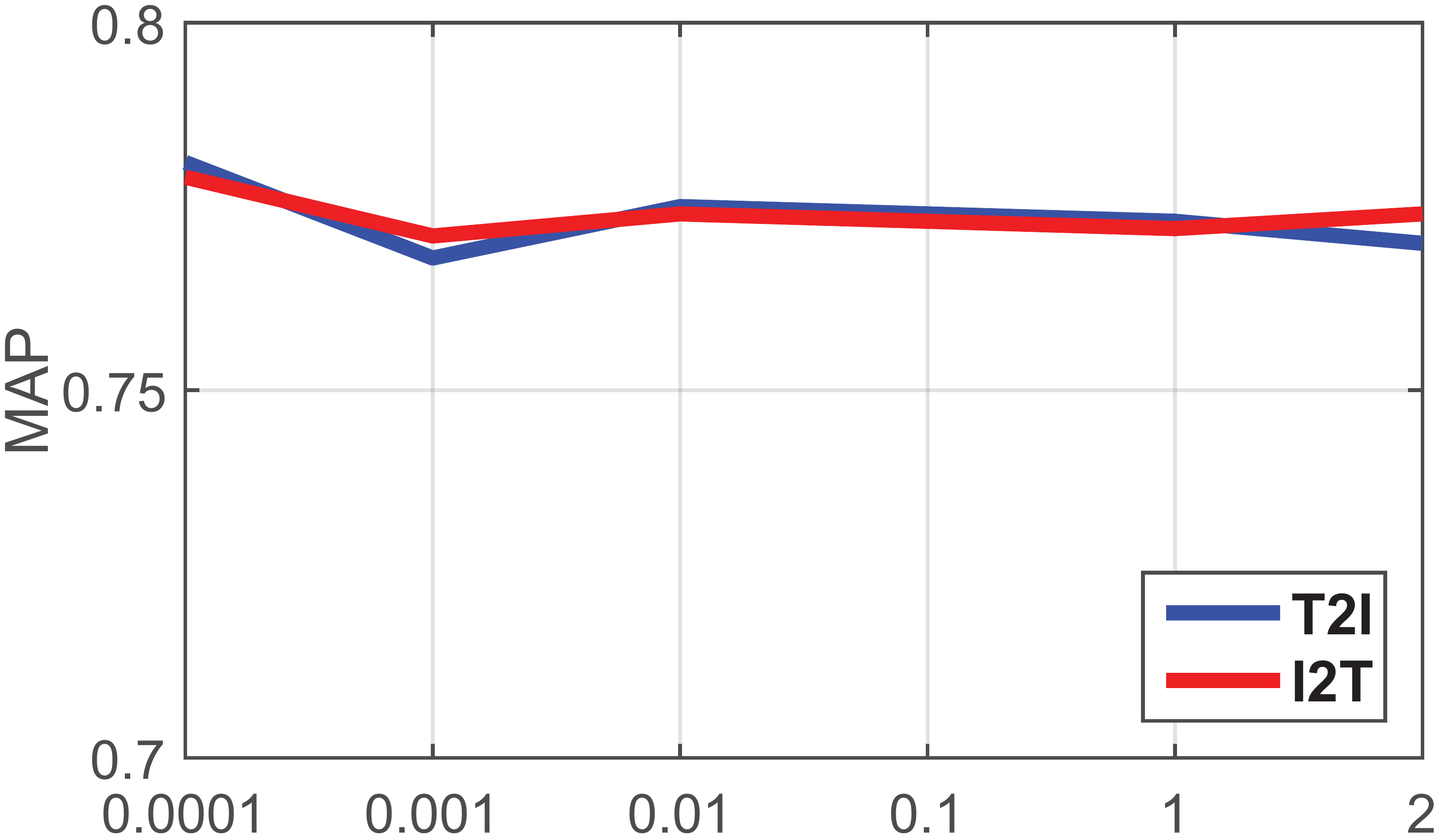}
		\vspace{-0.3cm}
		\caption*{(c) $\eta$}
	\end{minipage}
	\begin{minipage}[b]{0.24\textwidth}
		\vspace{-0.25cm}
		\centering
		\includegraphics[width=0.95\textwidth]{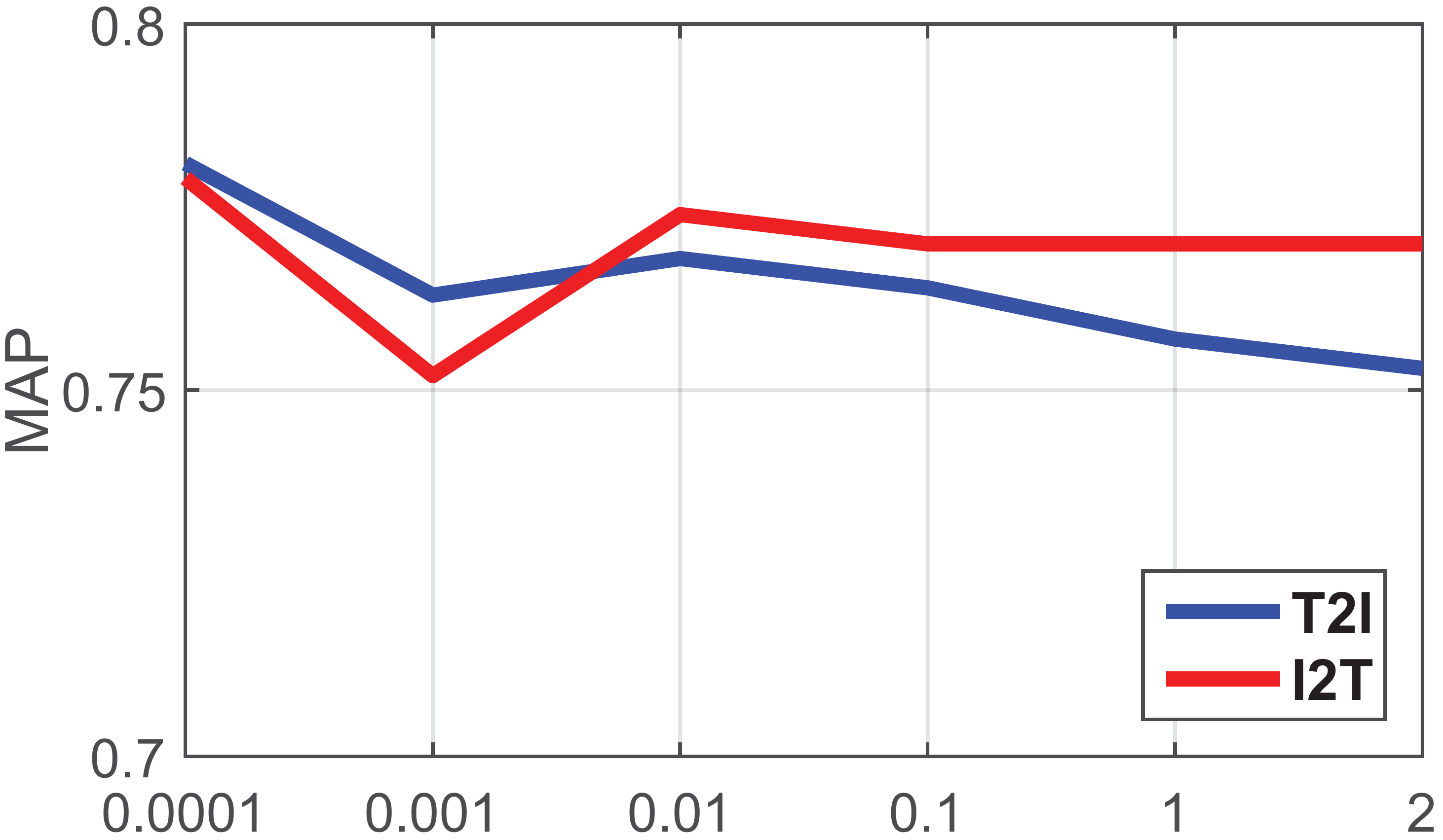}
		\vspace{-0.3cm}
		\caption*{(d) $\beta$}
	\end{minipage}
	\vspace{-0.25cm}
	\caption{A sensitivity analysis of the hyper-parameters.}
	\label{fig::hyperparameters}
	\vspace{-0.6cm}
\end{figure*}

In order to determine the hyper-parameters $\alpha$, $\gamma$, $\eta$, and $\beta$, we randomly select some data points (2,000 for each dataset) from the retrieval database to serve as our validation set. A sensitivity analysis of these hyper-parameters is provided in Fig.~\ref{fig::hyperparameters}. It can be seen that high performance can always be achieved when $\alpha$=$\gamma$=1 and $\eta$=$\beta$=$10^{-4}$. For image modality, we initialize the first seven layers of~\emph{ImgNet} with the CNN-F network pre-trained on the ImageNet dataset. For text modality,~\emph{TxtNet} randomly is initialized. The learning rate is chosen from $10^{-4}$ to $10^{-8}$. Following this, we show the average results of the 10 runs.
\vspace{-0.15cm}
\subsection{Performance}
\vspace{-0.2cm}
\textbf{Hamming Ranking:} Table \ref{resultAlex:Flickr-NUS-CoCo} reports the MAP results for both our SSAH and the other compared methods with CNN-F features on three popular datasets (MIRFLICKR-25K, NUS-WIDE and MS COCO) in cross-modal retrieval. ``I$\rightarrow$T'' denotes that the query is image and the database is text-based, and ``T$\rightarrow$I'' denotes that the query is text and the database is image-based. Compared with the shallow baselines of CVH, STMH, CMSSH, SCM and SePH, our SSAH achieves absolute more than a 10\% increase on MAP for I$\rightarrow$T/T$\rightarrow$I on the MIRFLICKR-25K dataset. While when comparing our SSAH with the deep-learning-based method (DCMH), we run the source code provided by the author. Here, it can be seen that SSAH can achieve more than a 5\% increase on MAP. For another two datasets NUS-WIDE and MS COCO with more instances and complex content, which are more challenging, SSAH always provides superior performance than other comparison methods, as presented in Table~\ref{resultAlex:Flickr-NUS-CoCo}. This may be because, during the learning process, the proposed self-supervised adversarial network more effectively facilitate the learning of semantic relevance between different modalities, which means that more discriminative representations can be learned using our SSAH. As a result, SSAH can more accurately capture correlations between modalities.

We further verify our SSAH using vgg19 features~\cite{simonyan2014very} that have been pre-trained on the ImageNet dataset. Table~\ref{resultVgg:Flickr-NUS-CoCo} shows the MAP results on three different datasets. As shown in Table~\ref{resultVgg:Flickr-NUS-CoCo}, we can see that almost all methods that are based on vgg19 outperform those based on CNN-F. Not only that, but it becomes evident that our SSAH consistently achieves the best performance. Compared with the shallow baselines (CVH, STMH, CMSSH, SCM and SePH),  SSAH achieves absolute more than 5\% increase on an average MAP for I$\rightarrow$T/T$\rightarrow$I on the MIRFLICKR-25K dataset. This means that the proposed SSAH can be applied to other networks and can achieve more accurate retrieval when equipped with an effective deep-network structure.
\begin{table*}[!t]		
	\vspace{-0.35cm}
	\begin{center}
		\caption{MAP. The best accuracy is shown in boldface. The baselines are based on CNN-F features.}
		\vspace{-0.3cm}
		\label{resultAlex:Flickr-NUS-CoCo}
		\begin{tabular}{|l|c|c|c|c|c|c|c|c|c|c|}
			\hline
			\multirow{2}{*}{TASK} & \multirow{2}{*}{Method} &\multicolumn{3}{c|}{Flickr-25K} & \multicolumn{3}{c|}{NUS-WIDE}& \multicolumn{3}{c|}{MS COCO}\\
			\cline{3-11}
			& & 16 bits& 32 bits& 64 bits& 16 bits& 32 bits& 64 bits& 16 bits& 32 bits& 64 bits\\
			\hline
			\multirow{7}{*}{I$\rightarrowtail$ T}
			& CVH \cite{Kumar2011Learning} & 0.557 & 0.554 & 0.554 & 0.400 & 0.392 & 0.386 & 0.412 & 0.401 & 0.400\\
			& STMH \cite{Wang2015Semantic}   & 0.602 & 0.608 & 0.605 & 0.522 & 0.529 & 0.537 & 0.422 & 0.459 & 0.475\\
			& CMSSH \cite{Bronstein2010Data}                  & 0.585 & 0.584 & 0.572 & 0.511 & 0.506 & 0.493 & 0.512 & 0.495 & 0.482\\
			& SCM \cite{Zhang2014Large}      & 0.671 & 0.682 & 0.685 & 0.533 & 0.548 & 0.557 & 0.483 & 0.528 & 0.550\\
			& SePH \cite{Lin2015CVPR}        & 0.657 & 0.660 & 0.661 & 0.478 & 0.487 & 0.489 & 0.463 & 0.487 & 0.501\\
			& DCMH \cite{Jiang2017CVPR}      & 0.735 & 0.737 & 0.750 & 0.478 & 0.486 & 0.488 & 0.511 & 0.513 & 0.527\\
			& \textbf{OURS} & $\mathbf{0.782}$ & $\mathbf{0.790}$ & $\mathbf{0.800}$ & $\mathbf{0.642}$ & $\mathbf{0.636}$ & $\mathbf{0.639}$ & $\mathbf{0.550}$ & $\mathbf{0.558}$ & $\mathbf{0.557}$ \\
			\hline
			\multirow{7}{*}{T$\rightarrowtail$ I}
			& CVH \cite{Kumar2011Learning} & 0.557 & 0.554 & 0.554 & 0.372 & 0.366 & 0.363 & 0.367 & 0.359 & 0.357\\
			& STMH \cite{Wang2015Semantic}   & 0.600 & 0.606 & 0.608 & 0.496 & 0.529 & 0.532 & 0.431 & 0.461 & 0.476\\
			& CMSSH \cite{Bronstein2010Data} & 0.567 & 0.569 & 0.561 & 0.449 & 0.389 & 0.380 & 0.429 & 0.408 & 0.398\\
			& SCM \cite{Zhang2014Large}      & 0.697 & 0.707 & 0.713 & 0.463 & 0.462 & 0.471 & 0.465 & 0.521 & 0.548\\
			& SePH \cite{Lin2015CVPR}        & 0.648 & 0.652 & 0.654 & 0.449 & 0.454 & 0.458 & 0.449 & 0.474 & 0.499\\
			& DCMH \cite{Jiang2017CVPR}      & 0.763 & 0.764 & 0.775 & 0.638 & 0.651 & 0.657 & 0.501 & 0.503 & 0.505\\
			& \textbf{OURS} & $\mathbf{0.791}$ & $\mathbf{0.795}$ & $\mathbf{0.803}$ & $\mathbf{0.669}$ & $\mathbf{0.662}$ & $\mathbf{0.666}$ & $\mathbf{0.537}$ & $\mathbf{0.538}$ &$\mathbf{0.529}$ \\
			\hline
		\end{tabular}
	\end{center}
	\vspace{-0.55cm}
\end{table*}
\begin{table*}[!t]
	\begin{center}
		\caption{MAP. The best accuracy is shown in boldface. The baselines are based on vgg19 features.}
		\vspace{-0.3cm}
		\label{resultVgg:Flickr-NUS-CoCo}
		\begin{tabular}{|l|c|c|c|c|c|c|c|c|c|c|}
			\hline
			\multirow{2}{*}{TASK} & \multirow{2}{*}{Method} &\multicolumn{3}{c|}{Flickr-25K} & \multicolumn{3}{c|}{NUS-WIDE}& \multicolumn{3}{c|}{MS COCO}\\
			\cline{3-11}
			& & 16 bits& 32 bits& 64 bits& 16 bits& 32 bits& 64 bits& 16 bits& 32 bits & 64 bits\\
			\hline
			\multirow{7}{*}{I$\rightarrowtail$ T}
			& CVH \cite{Kumar2011Learning} & 0.557 & 0.554 & 0.554 & 0.405 & 0.397 & 0.391 & 0.441 & 0.428 & 0.402\\
			& STMH \cite{Wang2015Semantic}   & 0.591 & 0.606 & 0.613 & 0.471 & 0.516 & 0.549 & 0.445 & 0.482 & 0.502\\
			& CMSSH \cite{Bronstein2010Data} & 0.593 & 0.592 & 0.585 & 0.508 & 0.506 & 0.495 & 0.504 & 0.495 & 0.492\\
			& SCM  \cite{Zhang2014Large}     & 0.685 & 0.693 & 0.697 & 0.497 & 0.502 & 0.499 & 0.498 & 0.556 & 0.565\\
			& SePH \cite{Lin2015CVPR}        & 0.709 & 0.711 & 0.716 & 0.479 & 0.501 & 0.492 & 0.489 & 0.502 & 0.499\\
			& DCMH \cite{Jiang2017CVPR}      & 0.677 & 0.703 & 0.725 & 0.590 & 0.603 & 0.609 & 0.497 & 0.506 & 0.511\\
			& \textbf{OURS} & $\mathbf{0.797}$ & $\mathbf{0.809}$ & $\mathbf{0.810}$ & $\mathbf{0.636}$ & $\mathbf{0.636}$ & $\mathbf{0.637}$ & $\mathbf{0.550}$ & $\mathbf{0.577}$ & $\mathbf{0.576}$\\
			\hline
			\multirow{7}{*}{T$\rightarrowtail$ I}
			& CVH \cite{Kumar2011Learning} & 0.557 & 0.554 & 0.554 & 0.385 & 0.379 & 0.373 & 0.413 & 0.402 & 0.388\\
			& STMH \cite{Wang2015Semantic}   & 0.600 & 0.613 & 0.616 & 0.472 & 0.526 & 0.550 & 0.446 & 0.478 & 0.506\\
			& CMSSH \cite{Bronstein2010Data} & 0.585 & 0.570 & 0.569 & 0.377 & 0.389 & 0.388 & 0.417 & 0.420 & 0.416\\
			& SCM  \cite{Zhang2014Large}     & 0.707 & 0.714 & 0.719 & 0.567 & 0.583 & 0.597 & 0.492 & 0.556 & 0.568\\
			& SePH \cite{Lin2015CVPR}        & 0.722 & 0.723 & 0.727 & 0.487 & 0.493 & 0.488 & 0.485 & 0.495 & 0.485\\
			& DCMH \cite{Jiang2017CVPR}      & 0.705 & 0.717 & 0.724 & 0.620 & 0.634 & 0.643 & 0.507 & 0.520 & 0.527\\
			& \textbf{OURS} & $\mathbf{0.782}$ & $\mathbf{0.797}$ & $\mathbf{0.799}$ & $\mathbf{0.653}$ & $\mathbf{0.676}$ & $\mathbf{0.683}$ & $\mathbf{0.552}$ & $\mathbf{0.578}$ & $\mathbf{0.578}$\\
			\hline
		\end{tabular}
	\end{center}
	\vspace{-0.5cm}
\end{table*}

\begin{figure*}[!t]
	\centering
	\subfloat[]{
		\begin{minipage}[b]{.24\linewidth}
			\vspace{-0.15cm}
			\centering
			\includegraphics[width=.99\textwidth]{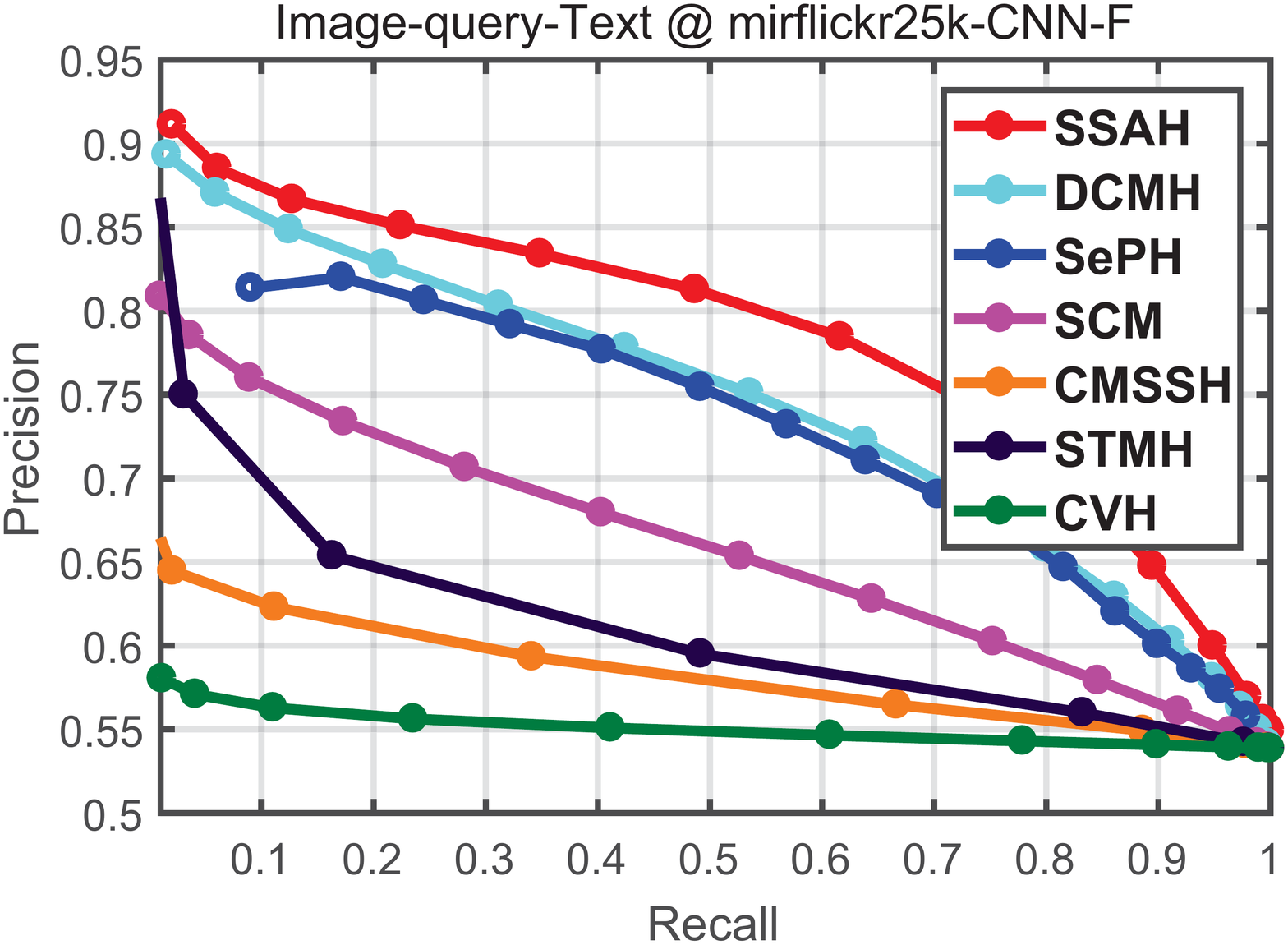}
	\end{minipage}}
	\subfloat[]{
		\begin{minipage}[b]{.24\linewidth}
			\vspace{-0.15cm}
			\centering
			\includegraphics[width=.99\textwidth]{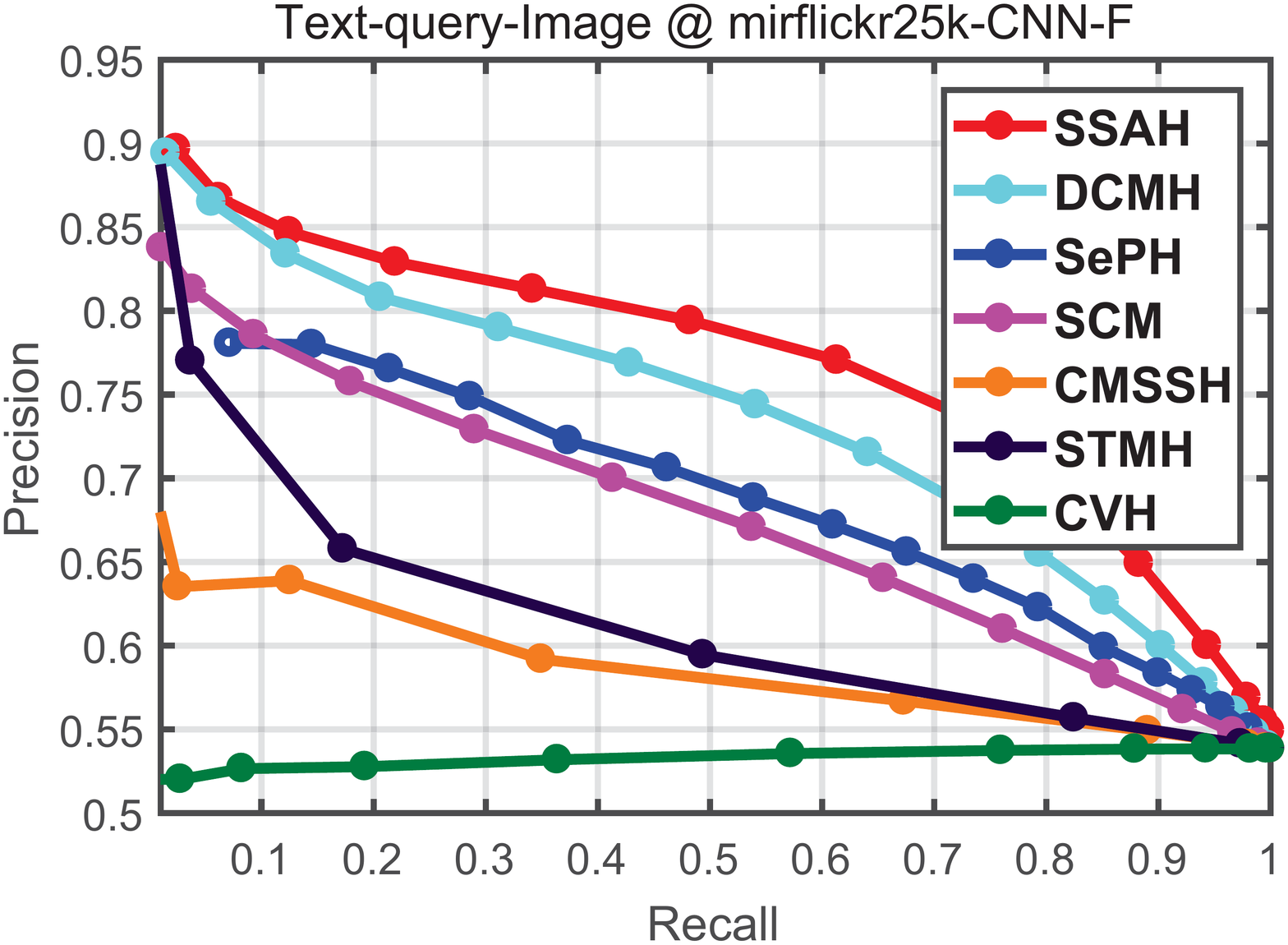}
	\end{minipage}}
	\subfloat[]{
		\begin{minipage}[b]{.24\linewidth}
			\vspace{-0.15cm}
			\centering
			\includegraphics[width=.99\textwidth]{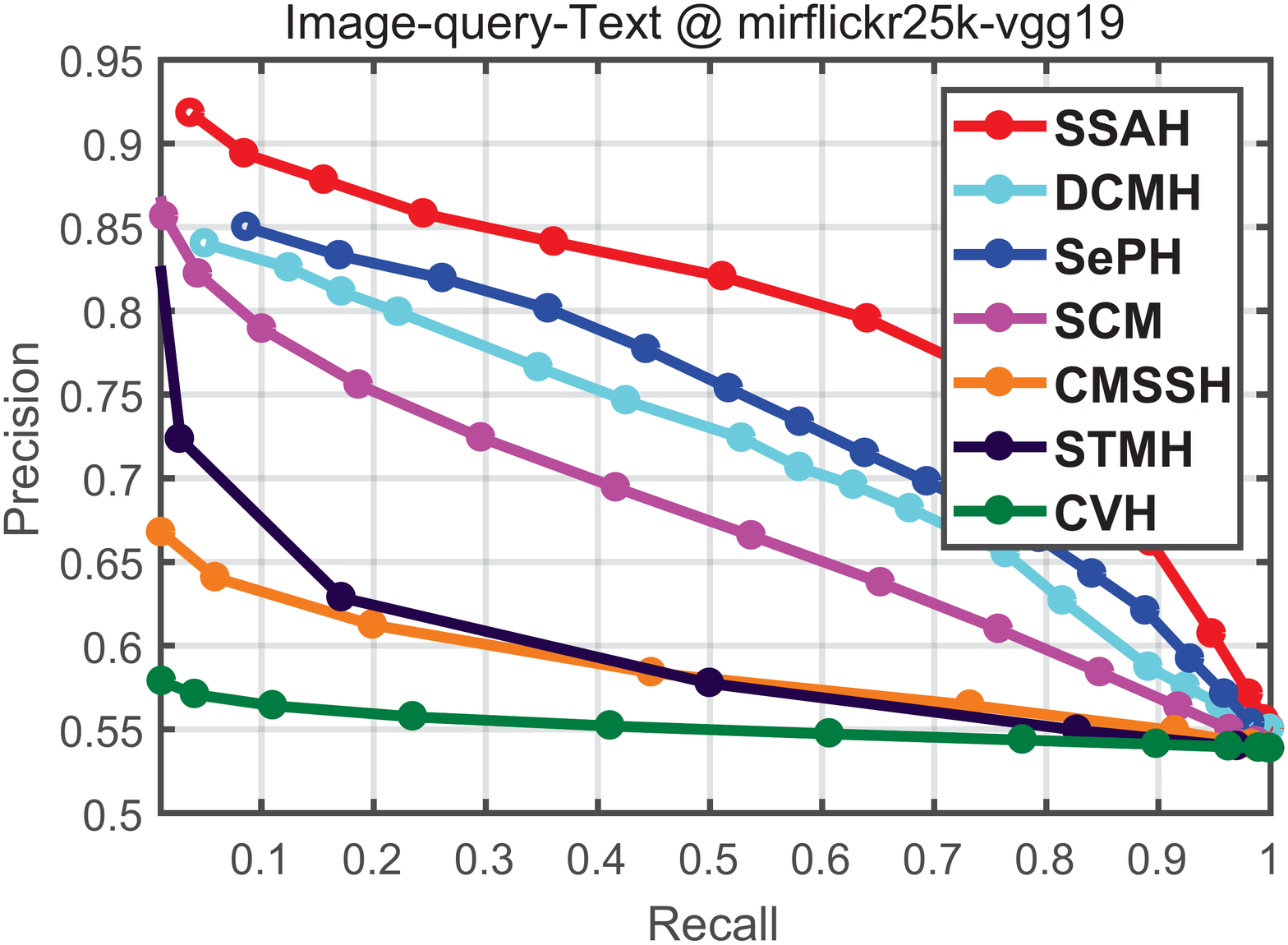}
	\end{minipage}}
	\subfloat[]{
		\begin{minipage}[b]{.24\linewidth}
			\vspace{-0.15cm}
			\centering
			\includegraphics[width=.99\textwidth]{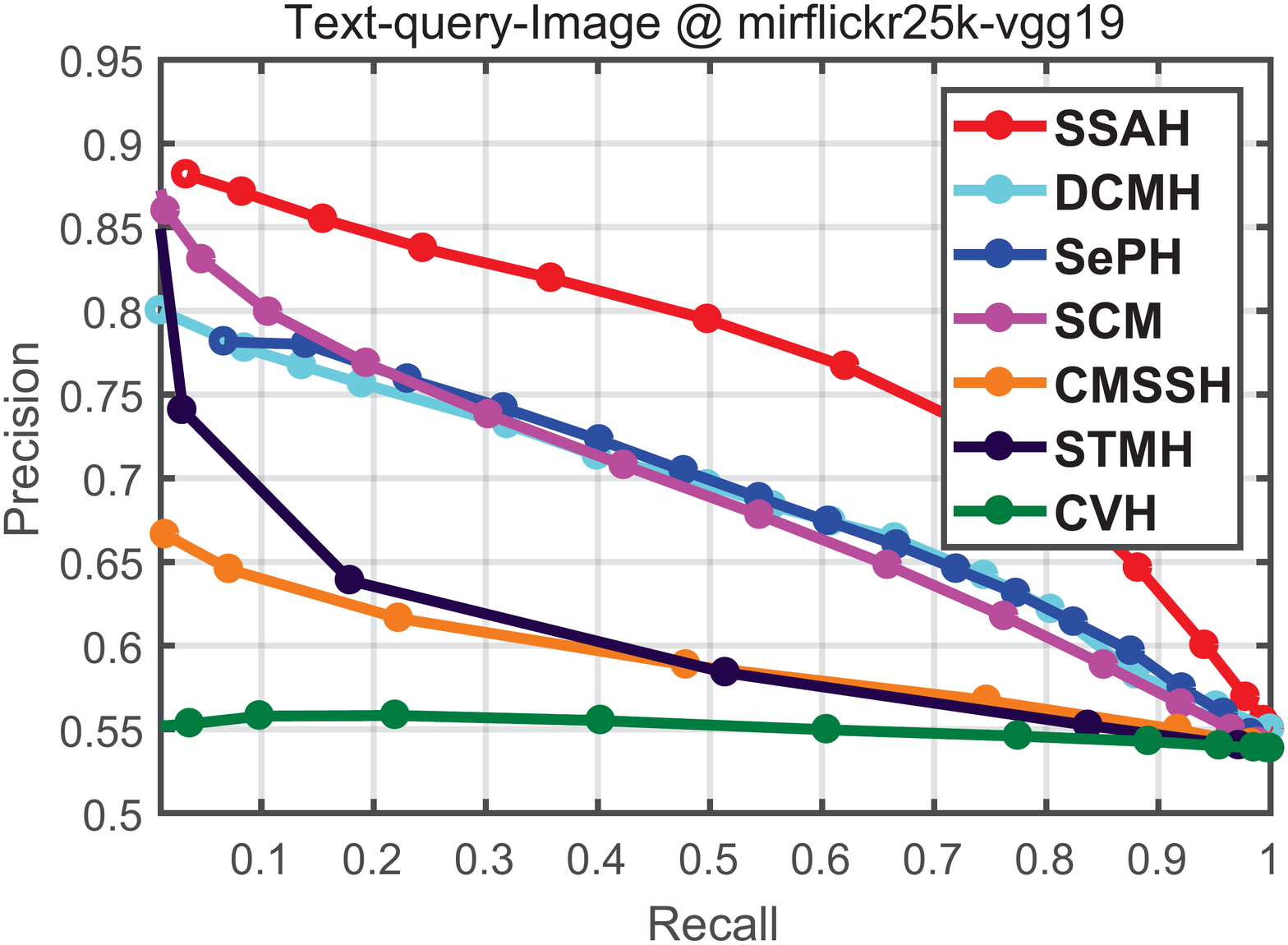}
	\end{minipage}}\\
	\vspace{-0.25cm}
	\subfloat[]{
		\begin{minipage}[b]{.24\linewidth}
			\vspace{-0.15cm}
			\centering
			\includegraphics[width=.99\textwidth]{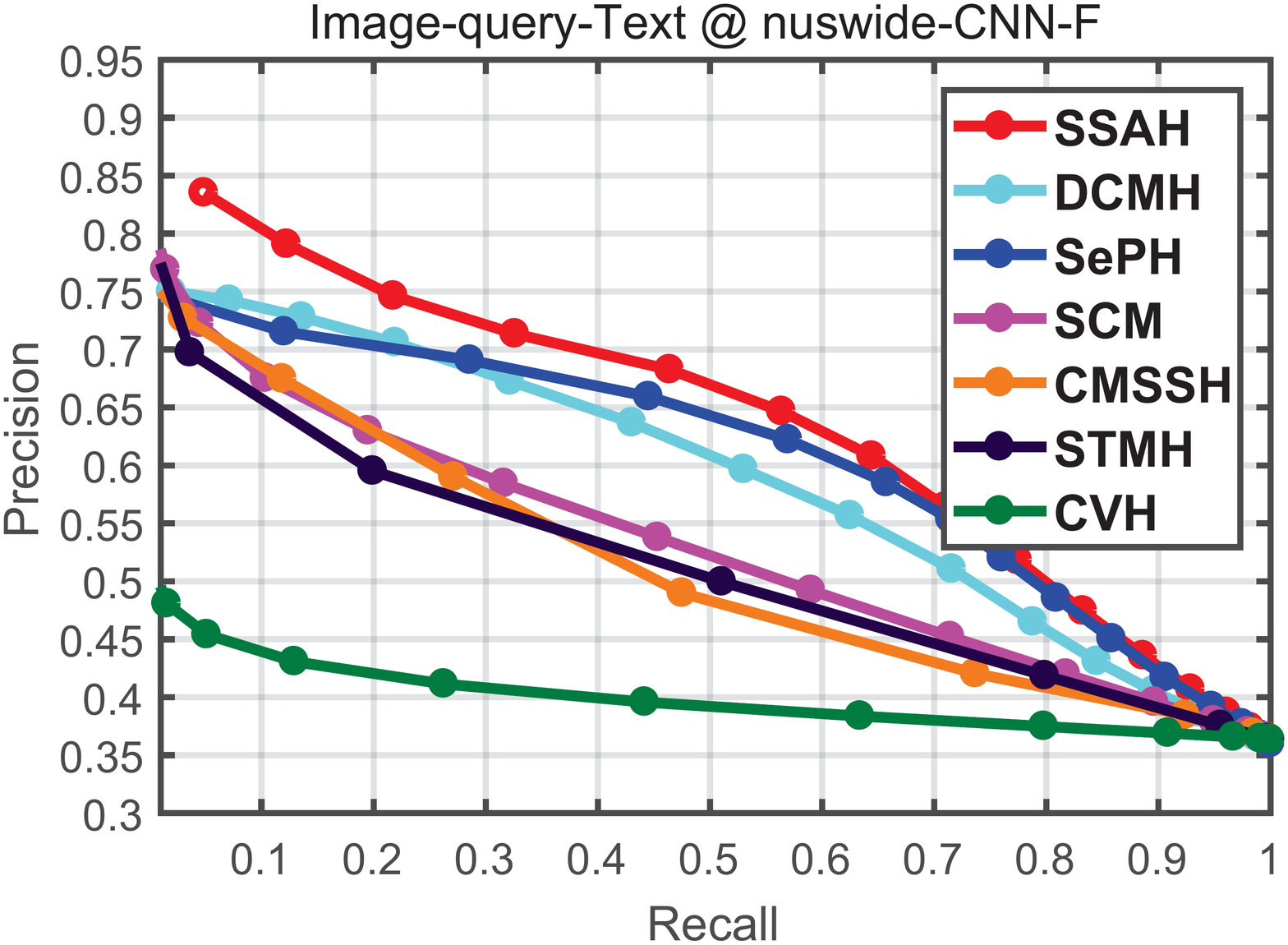}
	\end{minipage}}
	\subfloat[]{
		\begin{minipage}[b]{.24\linewidth}
			\vspace{-0.15cm}
			\centering
			\includegraphics[width=.99\textwidth]{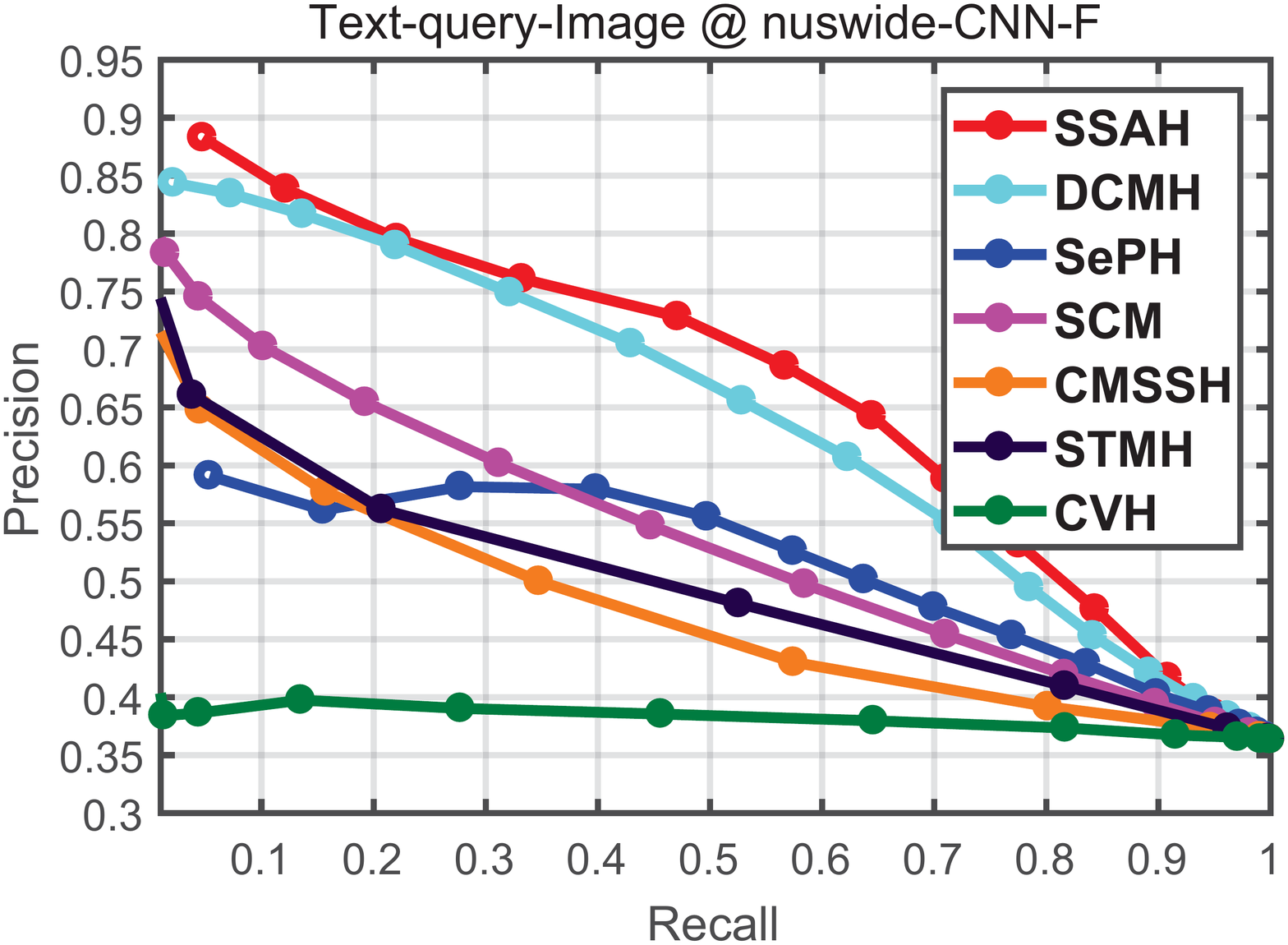}
	\end{minipage}}
	\subfloat[]{
		\begin{minipage}[b]{.24\linewidth}
			\vspace{-0.15cm}
			\centering
			\includegraphics[width=.99\textwidth]{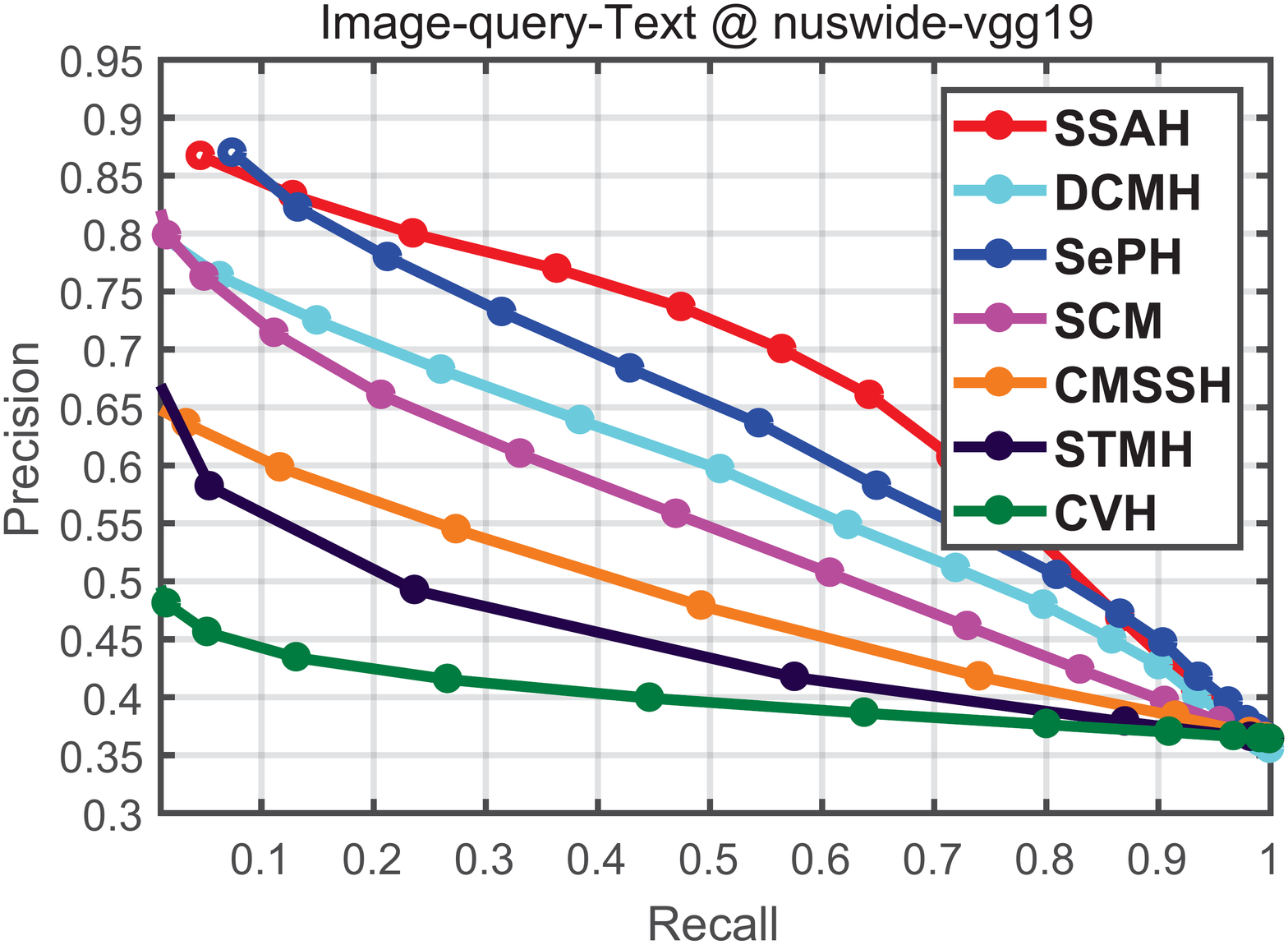}
	\end{minipage}}
	\subfloat[]{
		\begin{minipage}[b]{.24\linewidth}
			\vspace{-0.15cm}
			\centering
			\includegraphics[width=.99\textwidth]{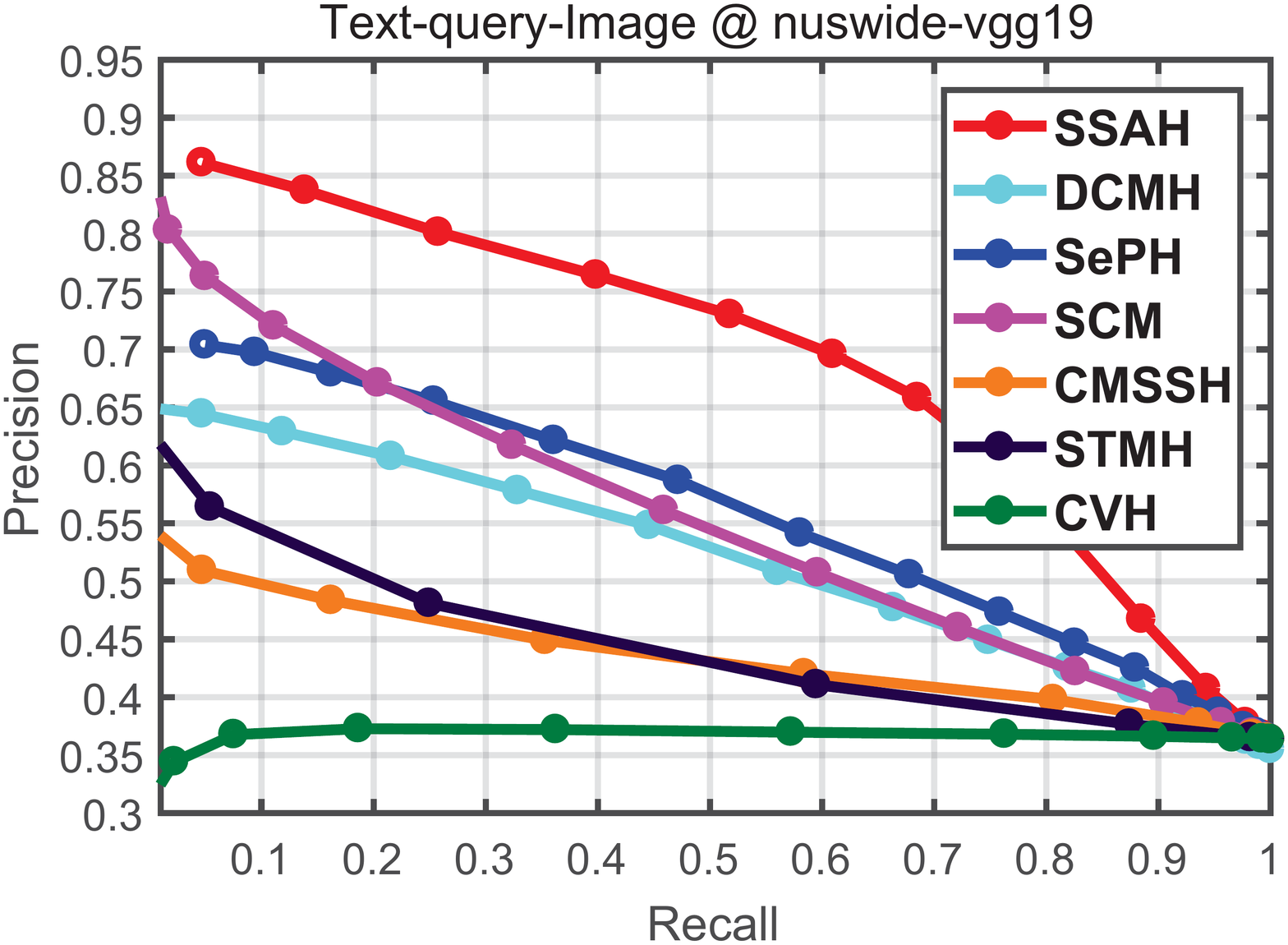}
	\end{minipage}}\\
	\vspace{-0.25cm}
	\subfloat[]{
		\begin{minipage}[b]{.24\linewidth}
			\vspace{-0.15cm}
			\centering
			\includegraphics[width=.99\textwidth]{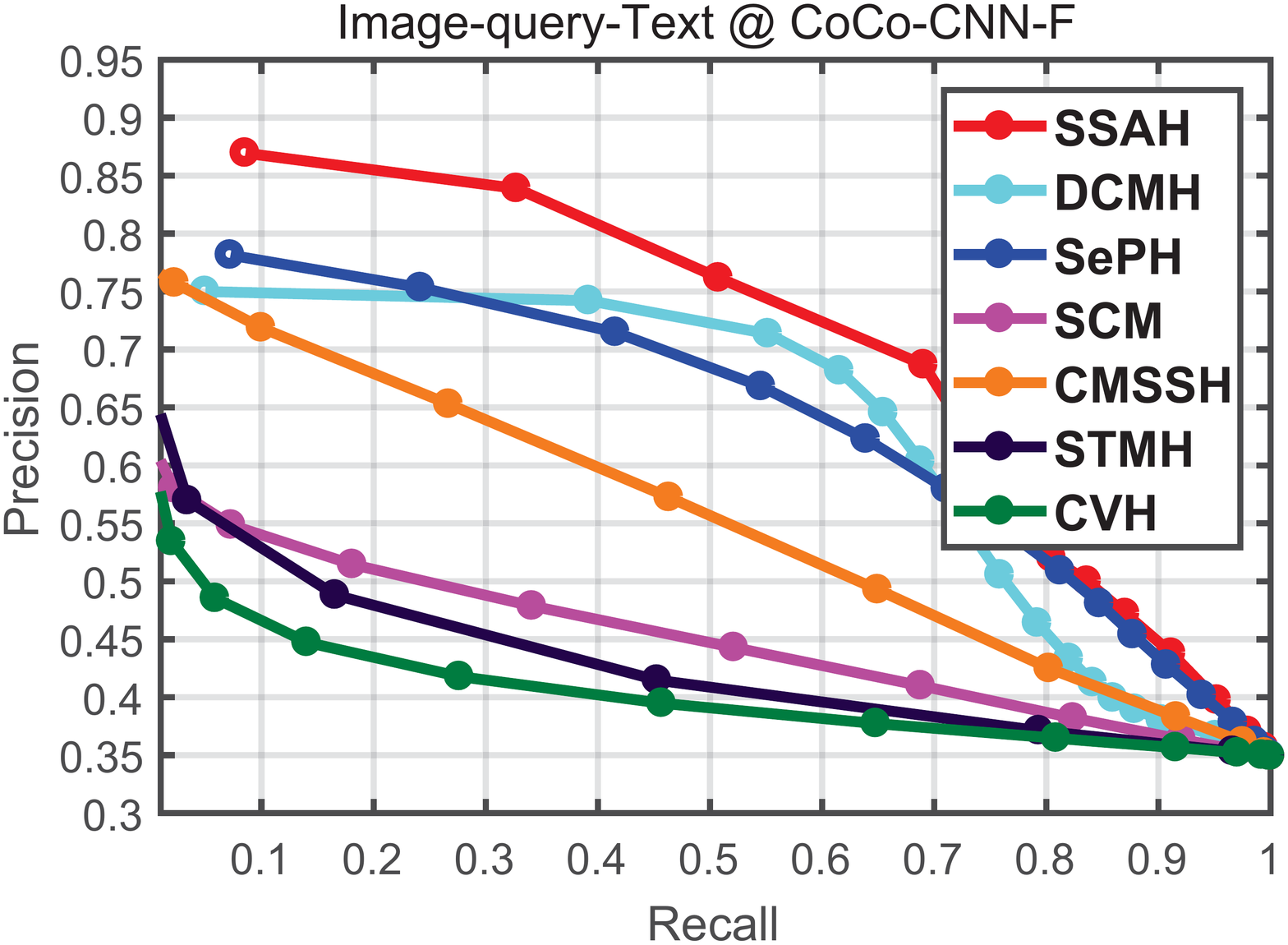}
	\end{minipage}}
	\subfloat[]{
		\begin{minipage}[b]{.24\linewidth}
			\vspace{-0.15cm}
			\centering
			\includegraphics[width=.99\textwidth]{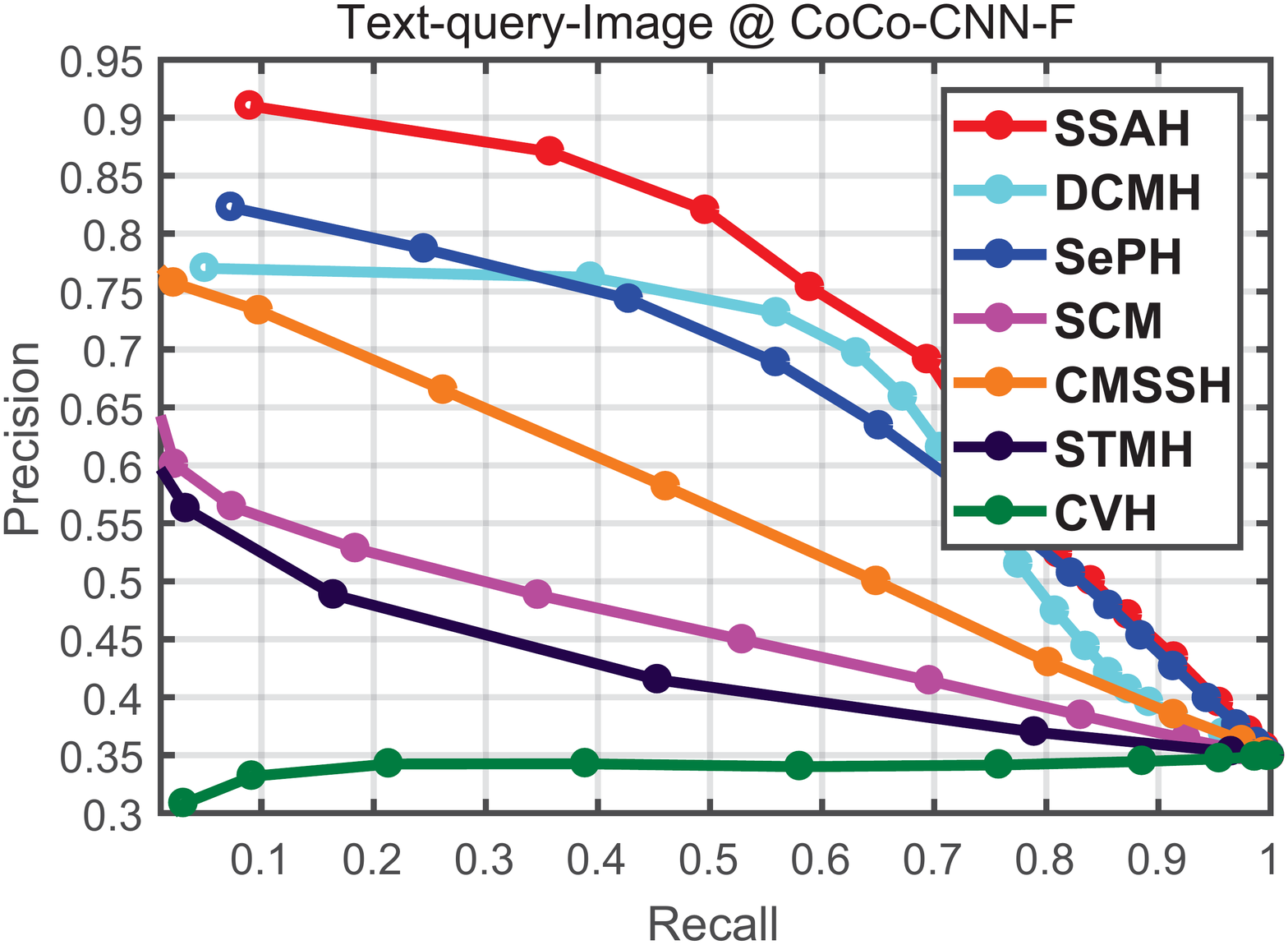}
	\end{minipage}}
	\subfloat[]{
		\begin{minipage}[b]{.24\linewidth}
			\vspace{-0.15cm}
			\centering
			\includegraphics[width=.99\textwidth]{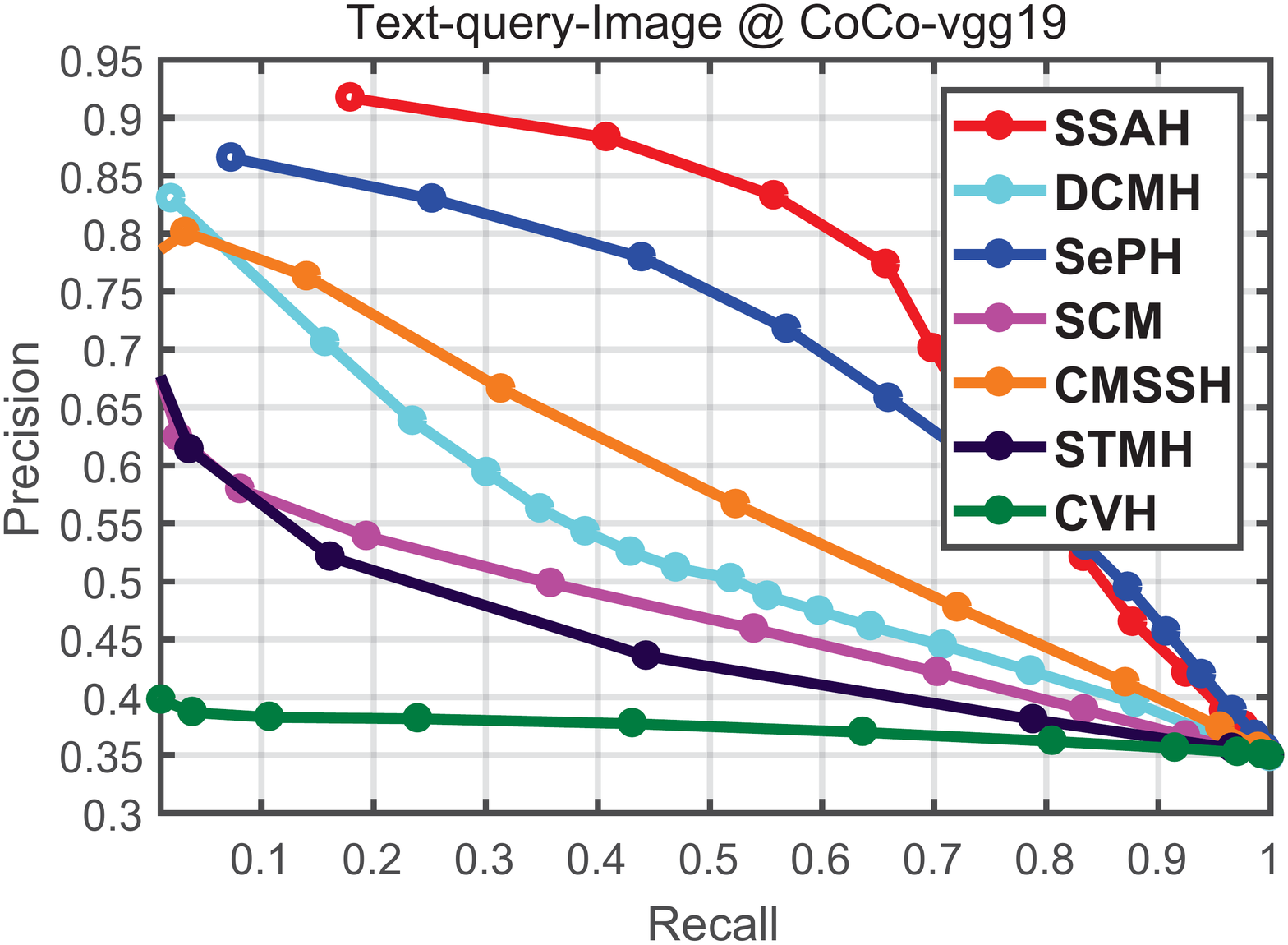}
	\end{minipage}}
	\subfloat[]{
		\begin{minipage}[b]{.24\linewidth}
			\vspace{-0.15cm}
			\centering
			\includegraphics[width=.99\textwidth]{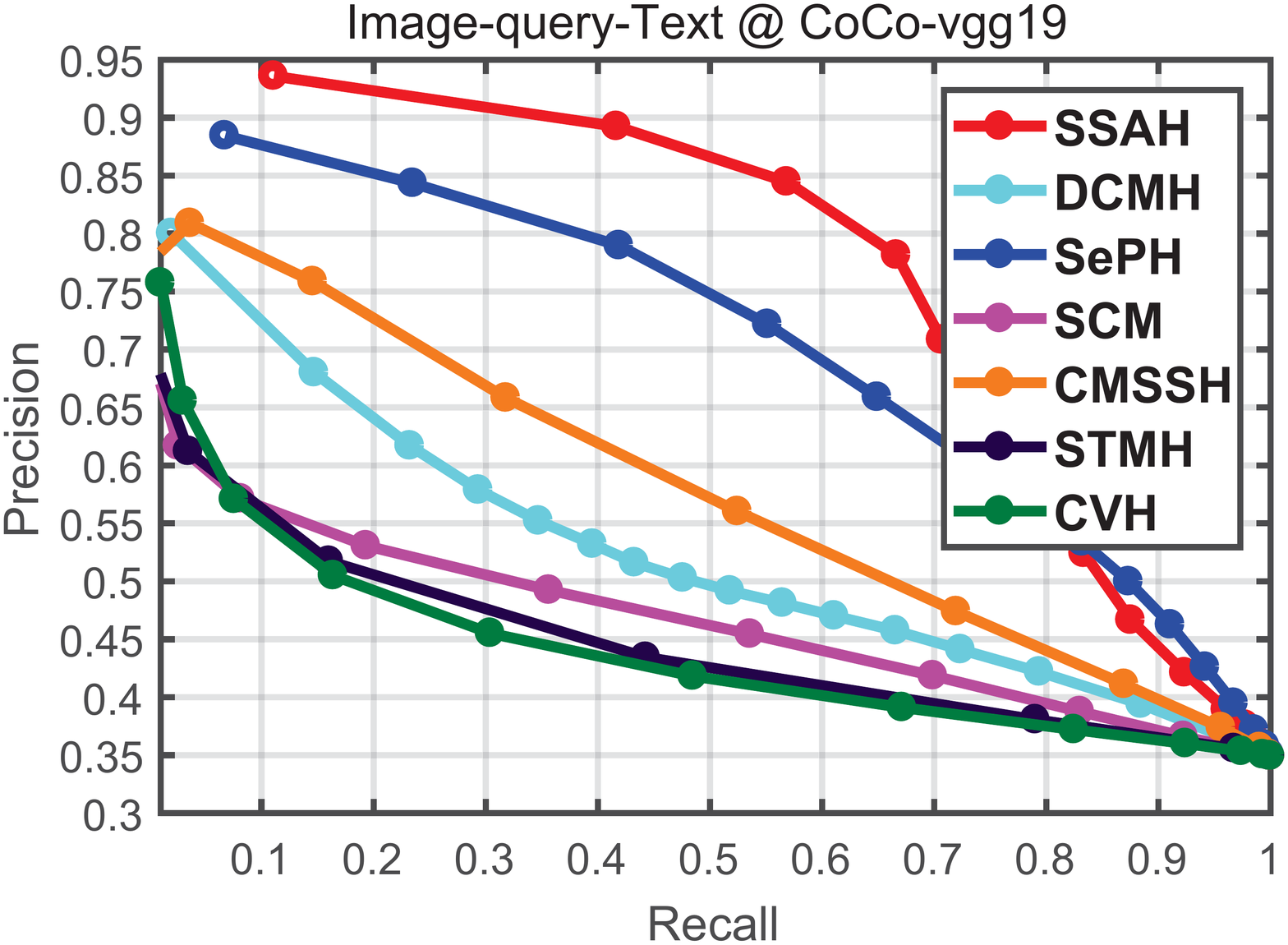}
	\end{minipage}}
	\vspace{-0.35cm}
	\caption{Precision-recall curves. The baselines are based on CNN-F features. The code length is 16.}
	\label{fig: PR curve}
	\vspace{-0.5cm}
\end{figure*}

\textbf{Hash Lookup:} When considering the lookup protocol, we compute the PR curve for the returned points given any Hamming radius. The PR curve can be obtained by varying the Hamming radius from $0$ to $16$ with a step-size of $1$. Fig.~\ref{fig: PR curve} shows the PR curves of all the current state-of-the-art methods with 16-bit hash codes on three benchmark datasets. In this way, it can be seen that our SSAH significantly outperforms all of its state-of-the-art competitors.
\begin{figure}[!t]
	\begin{center}
		\begin{minipage}[b]{0.24\textwidth}
			\centering
			\includegraphics[width=0.9\textwidth]{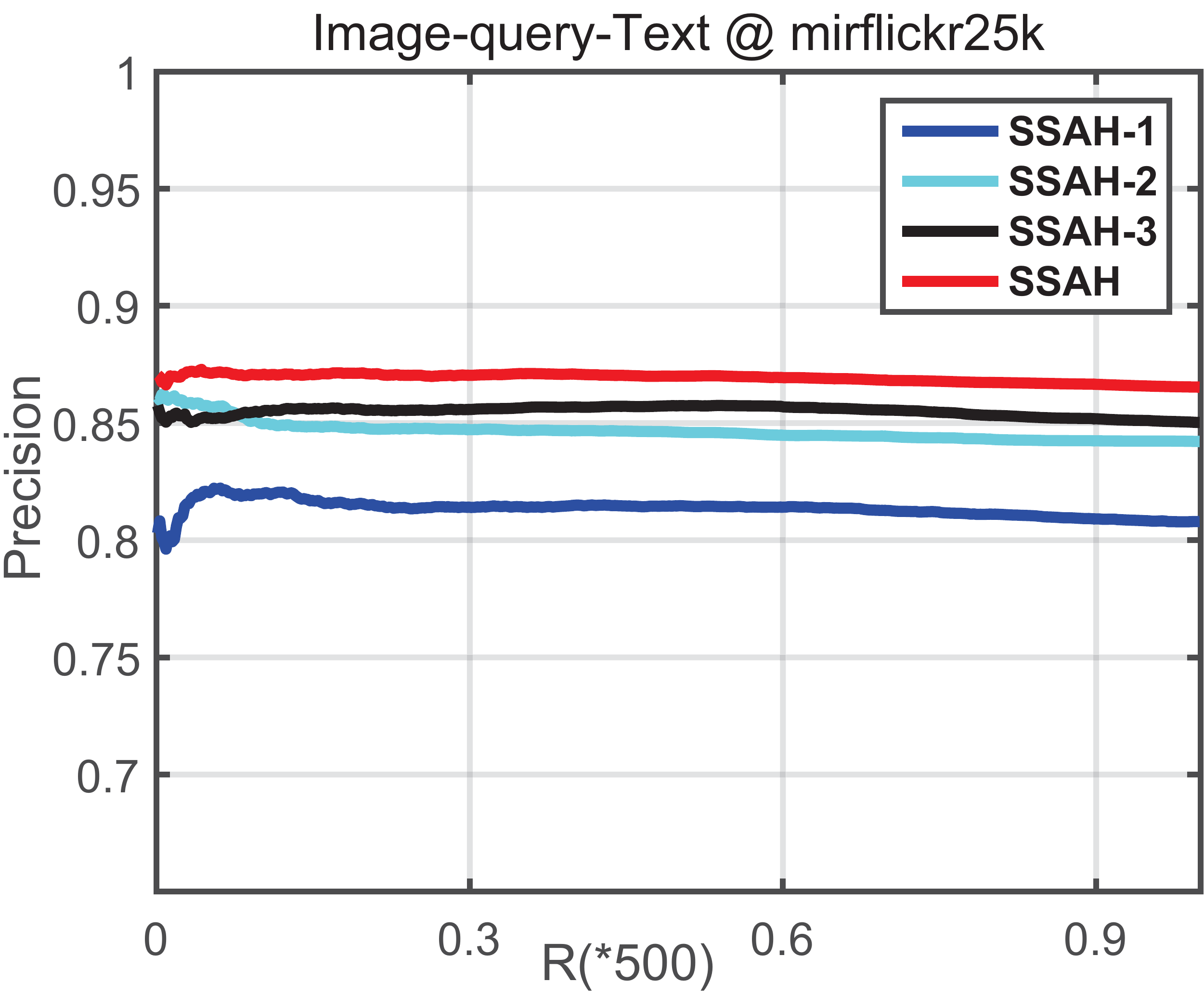}
			\vspace{-0.2cm}
			\caption*{(a) I2T@16 bit}
		\end{minipage}%
		\begin{minipage}[b]{0.24\textwidth}
			\centering
			\includegraphics[width=0.9\textwidth]{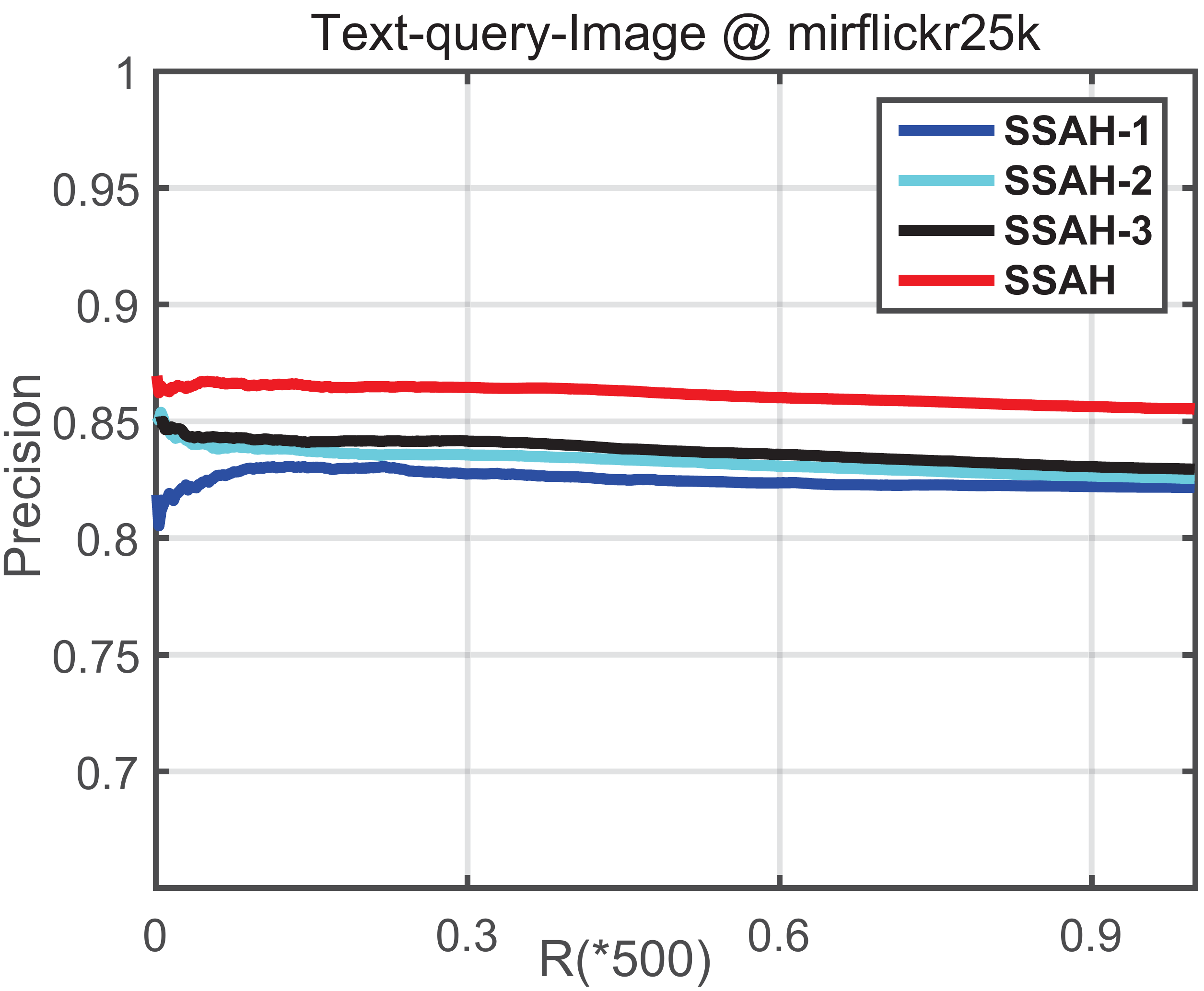}
			\vspace{-0.2cm}
			\caption*{(b) T2I@16 bit}
		\end{minipage}
	\end{center}
	\vspace{-0.55cm}
	\caption{Precision@top1000 curves on MIRFLICKR-25K.}
	\label{fig::MIRFLICKR-25K}
	\vspace{-0.5cm}
\end{figure}

\textbf{Ablation study of SSAH:} We also verify the impact of different network modules on our SSAH's performance. Three variants are designed as baselines of our SSAH networks: (a) SSAH-1 is built by removing the self-supervised semantic network; (b) SSAH-2 is built by replacing~\emph{TxtNet} with three full-connected layers; (c) SSAH-3 is built by removing the adversarial learning module. Fig.~\ref{fig::MIRFLICKR-25K} shows the comparison results with 16 bits on the MIRFLICKR-25K dataset. From the results, we can see that our method can achieve a more accurate performance when using the designed modules and that the self-supervised semantic network significantly improves the performance.

\begin{figure}[!t]
	\centering
	\begin{minipage}[b]{0.24\textwidth}
		\centering
		\includegraphics[width=0.95\textwidth]{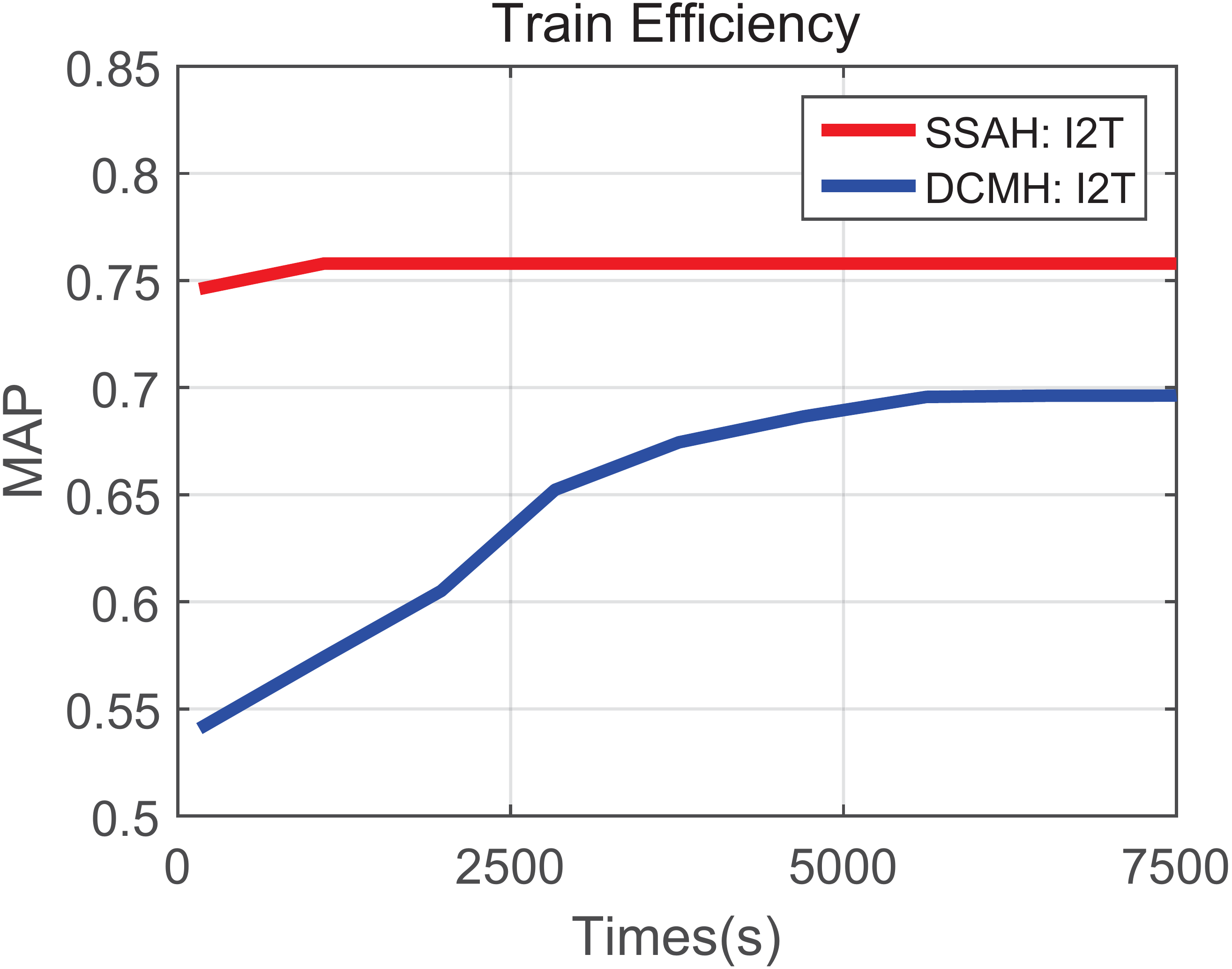}
		\vspace{-0.25cm}
		\caption*{(a) I2T@16 bit}
	\end{minipage}%
	\begin{minipage}[b]{0.24\textwidth}
		\centering
		\includegraphics[width=0.95\textwidth]{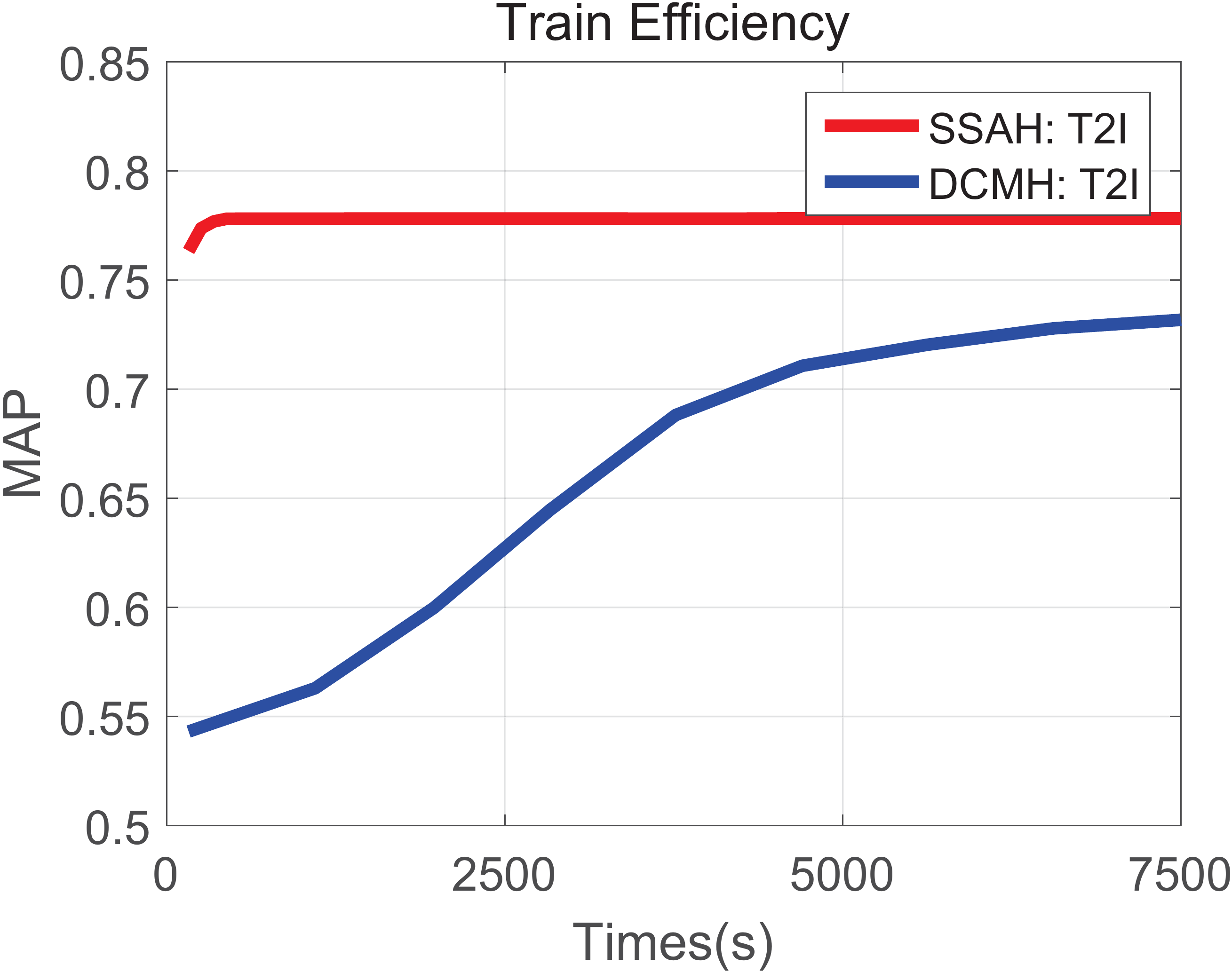}
		\vspace{-0.2cm}
		\caption*{(b) T2I@16 bit}
	\end{minipage}
	\vspace{-0.65cm}
	\caption{Training Efficiency of SSAH and DCMH.}
	\vspace{-0.7cm}
	\label{fig:DCMH_SSAH}
\end{figure}

\textbf{Training efficiency:} Fig.~\ref{fig:DCMH_SSAH} shows the variation between MAP and training time for SSAH and DCMH. We can see that our approach reduces training time by a factor of 10 over DCMH. In comparison to DCMH, SSAH exploits~\emph{LabNet} to learn more sufficient supervised information from high-dimensional semantic features and hash codes, using it to train~\emph{ImgNet} and~\emph{TxtNet} efficiently. Thus, more accurate correlations between different modalities can be captured and the modality gap can be  bridged more effectively.

\textbf{Comparison with ACMR:} According to our current best knowledge, ACMR~\cite{Wang2017MM} is the first work that borrows from adversarial learning approach for cross-modal retrieval. However, ACMR is not a hashing-based method. So as to be fairly compared with ACMR, we follow the experiment settings used in ACMR. SSAH is conducted on an NUS-WIDE-10k dataset, which is constructed by randomly selecting 10,000 image/text pairs from the 10 largest categories within the NUS-WIDE dataset. Table~\ref{result: NUS} shows the experiment results. The underlined results are reported in ACMR. It can be seen that our method outperforms ACMR significantly. This may be because two adversarial networks are used in our framework, with which SSAH can more accurately learn the distribution of different modalities and can thus capture the correlation more effectively.
\begin{table}[!t]
	\begin{center}
		\caption{MAP with CNN-F features on NUS-WIDE.}
		\vspace{-0.25cm}
		\label{result: NUS}
		\begin{tabular}{|l|c|c|c|c|}
			\hline
			Method   &  \multicolumn{2}{c|}{ACMR} &  \multicolumn{2}{c|}{SSAH}\\
			\hline
			Task & I$\rightarrow$ T & T$\rightarrow$ I & I$\rightarrow$ T & T$\rightarrow$ I\\
			\hline
			MAP &   \underline{0.544}  & \underline{0.538} &   \textbf{0.617}     &  \textbf{0.641}\\
			\hline
		\end{tabular}
		\vspace{-0.8cm}
	\end{center}
\end{table}
\vspace{-0.25cm}
\section{Conclusion}
\vspace{-0.15cm}
\label{section: Conclusion}
In this work, we proposed a novel deep hashing approach, dubbed self-supervised adversarial hashing (SSAH), in order to address the problem of cross-modal retrieval more effectively. The proposed SSAH incorporates a self-supervised semantic network coupled with multi-label information, and carries out adversarial learning to maximize the semantic relevance and feature distribution consistency between different modalities. The extensive experiments show that SSAH achieves state-of-the-art retrieval performance on three benchmark datasets.
\vspace{-0.25cm}
\section{Acknowledgements}
\vspace{-0.15cm}
\label{section: Acknowledgements}
This work was supported by the National Natural Science Foundation of China under Grant 61572388 and Grant 61703327, the Key R\&D Program/The Key Industry Innovation Chain of Shaanxi under Grant 2017ZDCXL-GY-05-04-02 and Grant 2017ZDCXL-GY-05-04-02, and ARC FL-170100117, DP-180103424, DP-140102164, LP-150100671.

{\small
\bibliographystyle{ieee}
\bibliography{egbib}
}

\end{document}